%% file: main.tex
\documentclass[journal,10pt,twocolumn]{IEEETran}
\usepackage[utf8]{inputenc}

\usepackage{threeparttable}
\usepackage{cite}
\usepackage{amsmath,amssymb,amsfonts}
\usepackage{bm}
\usepackage{algorithmic}
\usepackage{graphicx}
\usepackage{textcomp}
\usepackage{indentfirst}
\usepackage{graphicx}
\usepackage{csquotes}
\usepackage{multirow}
\usepackage{tabularx}
\usepackage{float}
\usepackage{color}
\usepackage{framed}
\usepackage{comment}
\usepackage{caption2}
\usepackage[ruled,linesnumbered]{algorithm2e}
\usepackage{upgreek}
\usepackage{soul}
\usepackage{siunitx}

\usepackage{tikz}

\SetKwProg{Fn}{Function}{}{}

\renewcommand{\figurename}{Figure}
\def\tablename{Table}

\def\BibTeX{{\rm B\kern-.05em{\sc i\kern-.025em b}\kern-.08em
    T\kern-.1667em\lower.7ex\hbox{E}\kern-.125emX}}

\graphicspath{{figures/}}

\newcommand*{\circled}[1]{\lower.7ex\hbox{\tikz\draw (0pt, 0pt)%
    circle (.5em) node {\makebox[1em][c]{\small #1}};}}

\begin{document}

\title{H2Learn: High-Efficiency Learning Accelerator for High-Accuracy Spiking Neural Networks}

\author{Ling Liang, Zheng Qu, Zhaodong Chen, Fengbin Tu, Yujie Wu, Lei Deng, \IEEEmembership{Member}, \IEEEmembership{IEEE}, Guoqi Li, \IEEEmembership{Member}, \IEEEmembership{IEEE}, Peng Li,  \IEEEmembership{Fellow}, \IEEEmembership{IEEE}, Yuan Xie, \IEEEmembership{Fellow}, \IEEEmembership{IEEE}\\

\thanks{Ling Liang, Zheng Qu, Zhaodong Chen, Fengbin Tu, Peng Li, and Yuan Xie are with the Department of Electrical and Computer Engineering, University of California, Santa Barbara, CA 93106, USA (email: \{lingliang, zhengqu, chenzd15thu, fengbintu, lip,  yuanxie\}@ucsb.edu). Yujie Wu, Lei Deng, Guoqi Li are with the Department of Precision Instrument, Center for Brain Inspired Computing Research, Tsinghua University, Beijing 100084, China (email: \{wu-yj16, leideng, liguoqi\}@mail.tsinghua.edu.cn). } }

\maketitle
 
\begin{abstract}

Although spiking neural networks (SNNs) take benefits from the bio-plausible neural modeling, the low accuracy under the common local synaptic plasticity learning rules limits their application in many practical tasks. Recently, an emerging SNN  supervised learning algorithm inspired by backpropagation through time (BPTT) from the domain of artificial neural networks (ANNs) has successfully boosted the accuracy of SNNs,  and helped improve the practicability of SNNs. However, current general-purpose processors suffer from low efficiency when performing BPTT for SNNs due to the ANN-tailored optimization. On the other hand, current neuromorphic chips cannot support BPTT because they mainly adopt local synaptic plasticity rules for simplified implementation. 

In this work, we propose \emph{H2Learn}, a novel architecture that can achieve high efficiency for BPTT-based SNN learning which ensures high accuracy of SNNs. At the beginning, we characterized the behaviors of BPTT-based SNN learning. Benefited from the binary spike based computation in the forward pass and the weight update, we first design look up table (LUT) based processing elements in Forward Engine and Weight Update Engine to make accumulations implicit and to fuse the computations of multiple input points. Second, benefited from the rich sparsity in the backward pass, we design a dual-sparsity-aware Backward Engine which exploits both input and output sparsity. Finally, we apply a pipeline optimization between different engines to build an end-to-end solution for the BPTT-based SNN learning. Compared with the modern NVIDIA V100 GPU, \emph{H2Learn} achieves $7.38\times$ area saving, $5.74\text{-}10.20\times$ speedup, and $5.25\text{-}7.12\times$ energy saving on several benchmark datasets.

\end{abstract}

\input{text/intro.tex}
\input{text/preliminary.tex}

\input{text/motivation.tex}
\input{text/architecture.tex}
\input{text/result.tex}

\input{text/related.tex}

\input{text/conclusion.tex}

\bibliography{./ref/ref}

\end{document}

%% file: text/intro.tex
\section{Introduction}

While the research of artificial neural networks (ANNs) such as deep neural networks (DNNs) have enjoyed great success in the past years \cite{lecun2015deep, he2016deep, tai2015improved,vaswani2017attention, mnih2015human}, extensive research of spiking neural networks (SNNs) are motivated by the bio-plausible neuron modeling, based on the observations that neurons use spike signals to represent information and communicate with each other.
Researchers have provided evidences that SNNs have unique advantages in processing naturally sparse and noisy information \cite{he2020comparing, liang2020exploring}.

How to train an SNN model with expected functionality is an essential problem for the SNN community. Many early studies have proposed unsupervised local learning based on the biological observation of local synaptic plasticity. In this family, spike timing dependent plasticity (STDP) \cite{song2000competitive, diehl2015unsupervised, Tavanaei2016Bio, Kheradpisheh2016STDP, liu2018brain, Lee2019Deep} has been widely explored, wherein each synaptic weight is modified locally based on the local spiking timing of the neurons wired by the synapse. However, such local synaptic plasticity suffers low accuracy and limited model scale, that is why its use in practical applications has been limited.

In order to improve the accuracy of SNNs, the algorithms in training ANNs are borrowed. For example, the ANN-to-SNN-conversion learning \cite{diehl2015fast, hu2018spiking, sengupta2019going} converts a trained ANN model into its SNN counterpart during inference. Although this method significantly improves accuracy of the resulting SNN, 
it needs to introduce an extra model switch and results in a long time duration to maintain accuracy,
which is not friendly for hardware implementation. Recently, an explicit format of gradient descent to train ANNs has been adapted and applied in training SNNs \cite{lee2016training}. Due to the spatio-temporal data paths in SNNs, backpropagation through time (BPTT) is a good fit. Previous studies in the algorithm level have demonstrated the effectiveness of BPTT for SNN learning \cite{wu2018spatio, jin2018hybrid, bellec2018long, Shrestha2018SLAYER, wu2019direct, gu2019stca}, which can achieve high accuracy and eliminate the issues of extra model switch and long time duration in the ANN-to-SNN-conversion learning.

Besides the functionality, how to train SNNs efficiently is also an important research topic. Currently, GPUs are still the mainstream platform for neural network training, while they are tailored for ANNs rather than SNNs. This can be reflected by the ANN-aware optimization for the GPUs' hardware architectures, programming libraries, training frameworks, etc. However, such optimization cannot fully utilize the special data format and computing paradigm of SNNs, thus causing inefficiencies when training SNNs on GPUs.

Beyond GPUs, researchers have also developed domain-specific chips for SNNs, usually termed as neuromorphic chips \cite{schemmel2010wafer, jin2010implementing, qiao2015reconfigurable, frenkel20180, frenkel2019morphic, davies2018loihi, baek2019flexlearn}. Here we focus on the ones targeting SNN learning rather than inference \cite{benjamin2014neurogrid, merolla2014million, moradi2017scalable, pei2019towards, deng2020tianjic, narayananspinalflow}. Nearly all currently available SNN learning chips adopt local synaptic plasticity such as STDP for weight update. The good locality without backpropagation makes it easier to implement on decentralized many-core neuromorphic architectures. Although they enjoy low power and fast response, it still cannot escape from the low accuracy of these local learning rules. This is also one of the major reasons why neuromorphic chips have not yet achieved the similar commercial success as deep learning accelerators.

In this work, 
we propose \emph{H2Learn}, a specific architecture for high-efficiency BPTT-based SNN learning. 
After a detailed profiling, 
we find special characteristics of BPTT in the context of SNNs, which are different from those for recurrent ANNs. Each learning iteration includes a forward pass, a backward pass, and a weight update stage. One of the two operands for the computational operations in the forward pass and weight update is the binary spike (0 or 1); moreover, the inputs and outputs of the computational operations in the backward pass are quite sparse. For the former feature, we propose a look up table (LUT) based processing element (PE) design that directly stores the partial sums of the results in the LUT and uses the input spikes as addresses to access them. 
For the latter feature, we design a dual-sparsity-aware architecture and compress the membrane potentials in memory. In this way, the redundant computation and storage can be significantly reduced. Finally, an overall pipeline to optimize the entire learning process is implemented. We summarize our contributions as follows:

\begin{itemize}

\item We identify opportunities to achieve both high accuracy and high efficiency in the BPTT algorithm for SNN learning, which overcomes the low accuracy of local synaptic plasticity and the high cost of backpropagation.

\item We propose the \emph{H2Learn} architecture for SNN learning, which consists of a Forward Engine, a Backward Engine, and a Weight Update Engine to execute the three phases in BPTT. In Forward Engine and Weight Update Engine, we design LUT-based PEs to make accumulations implicit and fuse the computations of multiple input points; in Backward Engine, we design an architecture that is aware of both input and output sparsity with less compute overhead and compress the membrane potentials with less memory overhead. The pipeline between different engines is also elaborated to improve the overall performance.

\item We conduct extensive experimental evaluations. Our LUT-based Forward Engine/Weight Update Engine and dual-sparsity-aware Backward Engine achieve significant superiority compared with the non-LUT and dense baselines, respectively. On several widely used benchmark datasets, we achieve significantly higher accuracy than prior neuromorphic chips that use local synaptic plasticity; furthermore, \emph{H2Learn} demonstrates 7.38$\times$ area saving, 5.74-10.20$\times$ speedup, and 5.25-7.12$\times$ energy saving compared with NVIDIA V100 GPU.
\end{itemize}

%% file: text/preliminary.tex
\section{Preliminaries of SNNs}

SNNs represent the neural networks inspired by the behaviors of neural circuits in the brain. Leaky integrate-and-fire (LIF) is the most widely used spiking neuron model which can capture the typical neuronal behaviors and easy for implementation in hardware. In this work, we focus on the learning of LIF-based SNNs.

\begin{figure}[!htbp]
\centering
\includegraphics[width=0.325\textwidth]{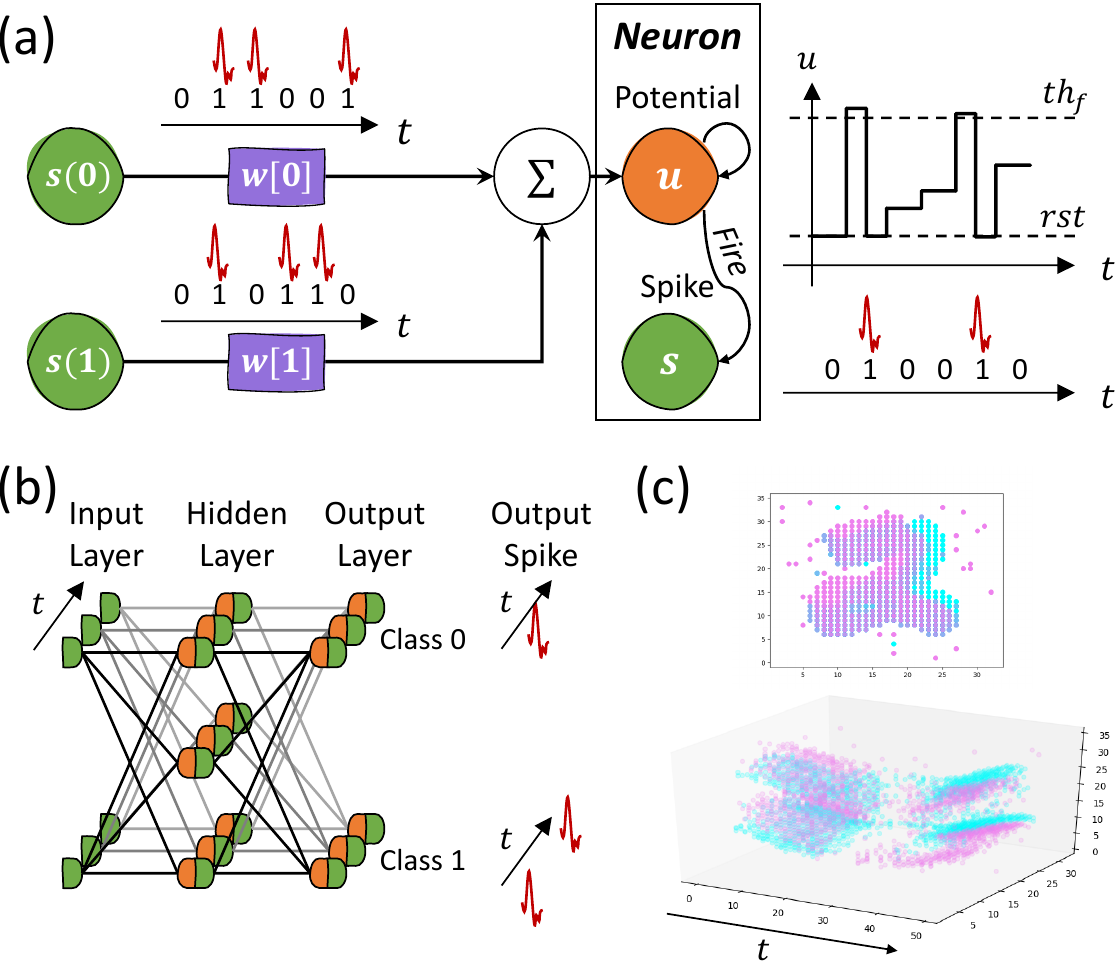}
\caption{Introduction of SNNs: (a) behaviors of a single spiking neuron; (b) a spiking network; (c) the input format.}
\label{fig:SNN}
\end{figure}

Fig. \ref{fig:SNN}(a) illustrates the behaviors of a spiking neuron. The inputs are first weighted by synapses and then integrates by dendrites to update the state of membrane potential (termed as potential in the rest of this work for simplicity) at soma. Once the potential exceeds a threshold ($th_f$), the neuron fires a spike event to its post-connected neurons and resets its potential to a reset value (usually zero); otherwise, nothing happens but the leakage of the potential. A spiking neuron in SNNs is different from an artificial neuron in ANNs. Specifically, (1) there is an intrinsic temporal domain in a spiking neuron but not in an artificial neuron; (2) 
the potential updating of a spiking neuron depends on both the historic state and the input integration, while the accumulated pre-activation in an artificial neuron just integrates inputs; (3) spiking neurons communicate with each other using binary spike events (0 or 1) while artificial neurons use continuous activations. 

Fig. \ref{fig:SNN}(b) shows a spiking network. The information propagates in both spatial and temporal domains. Due to the extra temporal domain compared with ANNs, the input format for SNNs is usually a 3D spike pattern rather than a 2D image, as shown in Fig. \ref{fig:SNN}(c). 
The output of an SNN is in a 2D spike pattern rather than a 1D vector in ANNs. The classification result is determined by the coding scheme of output. The rate coding is the commonly adopted one that the neuron fires the most indicates the recognized class. Notice that the network structure of SNNs can be arbitrary in principle, but the fully-connected (FC) layers based multilayered perceptron (MLP) and the convolutional (Conv) layers based convolutional neural network (CNN) are two usual cases, which is similar to ANNs.

%% file: text/motivation.tex
\section{Motivation}

Currently, two bottlenecks hinder the progress of SNNs: (1) low accuracy of conventional local learning rules (e.g., STDP), limiting their competitiveness and application scope in practice; (2) low execution efficiency on GPUs, limiting the exploration of the model scale and space (indirectly limiting accuracy). The former can be significantly improved by the BPTT algorithm, and the latter is due to GPUs’ specific optimization for ANNs, rather than for SNNs with special data format and computing paradigm. Therefore, we propose to design an efficient accelerator for BPTT-based SNN training to improve the competitiveness of neuromorphic chips. 

\subsection{Low-Accuracy SNN Learning on Neuromorphic Chips}

In order to build high-efficiency domain-specific chips for SNN learning, researchers have designed neuromorphic chips. However, almost all of them \cite{schemmel2010wafer, jin2010implementing, qiao2015reconfigurable, frenkel20180, frenkel2019morphic, davies2018loihi, baek2019flexlearn} adopt unsupervised learning rules inspired by bio-plausible synaptic plasticity, such as STDP \cite{song2000competitive}, the good locality of which makes it hardware friendly. 
However, in practical applications, this rule cannot be accepted due to the low accuracy when performing mainstream tasks (see Table \ref{tab:STDP_BPTT}) and the difficulty in scale-up when encounter complex tasks, which are the major reasons why neuromorphic chips are suffering skepticisms and are not applied widely as deep learning accelerators. 


\begin{table}[!htbp]
\caption{Accuracy comparison for SNN learning: STDP vs. BPTT.}
\label{tab:STDP_BPTT}
\vspace{5pt}
\centering
\renewcommand\arraystretch{1.1}
\resizebox{0.35\textwidth}{!}{
\begin{tabular}{c|c|c|c|c}
\hline \hline
\textbf{Method} & \textbf{Ref} & \textbf{Dataset} & \textbf{Net} & \textbf{Accuracy}\\
\hline
\multirow{5}*{STDP} & \cite{Lee2019Deep} & MNIST & CNN &  91.10\% \\
& \cite{Tavanaei2016Bio} & MNIST & CNN & 93.30\% \\
& \cite{diehl2015unsupervised} & MNIST & MLP & 95.00\% \\
& \cite{liu2018brain} & MNIST & CNN & 97.50\% \\
& \cite{Kheradpisheh2016STDP} & MNIST & CNN & 98.40\% \\
\hline
 \multirow{9}*{BPTT} &  \cite{gu2019stca} & MNIST & MLP & 98.60\% \\
& \cite{wu2018spatio} & MNIST & MLP & 98.89\% \\
& \cite{jin2018hybrid} & MNIST & MLP & 98.93\% \\
& \cite{jin2018hybrid} & MNIST & CNN & 99.49\% \\
& \cite{jin2018hybrid} & N-MNIST & MLP & 98.88\% \\
& \cite{Shrestha2018SLAYER} & N-MNIST & CNN & 99.20\% \\
 & \cite{wu2019direct} & N-MNIST & CNN & 99.44\% \\
 & \cite{wu2019direct} & CIFAR10 & CNN & 89.83\% \\
  & \cite{wu2019direct} & CIFAR10-DVS & CNN & 58.10\% \\
\hline \hline
\end{tabular}}

\end{table}

\subsection{Low-Efficiency SNN Learning on GPUs}

GPUs play the backbone role in neural network training. However, current GPU hardware architectures (e.g., tensor cores on NVIDIA GPUs), programming libraries (e.g., cuDNN), and model training frameworks (e.g., TensorFlow and Pytorch) are mainly optimized for ANNs rather than SNNs, causing inefficiencies when training SNN models.

Specifically, the inefficiencies come from several aspects. First, the neuronal activities in SNNs are binary spikes, while the computation and storage data formats on GPUs are floating points for the continuous activations in ANNs. 
Second, there is rich sparsity during backpropagation of SNN training, 
while GPUs prefer dense computation and storage. At last, the ANN-aware dataflow optimization on GPUs to reduce the memory footprint for intermediate data is not suited for SNNs due to the distinct model behaviors. Table \ref{tab:cnn_snn_compare} shows that SNN training is much slower than ANN training on GPU under the same network structure.

\begin{table}[!htbp]
\caption{Latency of one training epoch for ANNs and SNNs under the same network structure on NIVIDIA V100 GPU.}
\label{tab:cnn_snn_compare}
\vspace{5pt}
\centering
\renewcommand\arraystretch{1.1}
\resizebox{0.45\textwidth}{!}{
\begin{tabular}{c|c|c|c}
\hline \hline
\textbf{Dataset} & \textbf{Latency of ANNs} & \textbf{Latency of SNNs} & \textbf{Performance Drop}\\
\hline

MNIST   & 12.12s & 138.55s & 11.43$\times$ \\
\hline
CIFAR10 & 12.98s & 147.07s & 11.33$\times$ \\
\hline
ImageNet & 0.72hr & 4.00hr & 5.56$\times$ \\

\hline \hline
\end{tabular}}

\end{table}

\subsection{High-Accuracy and High-Efficiency SNN Learning}

\textbf{BPTT Learning for High Accuracy.} Recently, researchers began to borrow ideas from the learning of ANNs. Typically, the gradient-descent-based backpropagation algorithms have been applied in SNN training \cite{lee2016training, wu2018spatio, jin2018hybrid, bellec2018long, Shrestha2018SLAYER, wu2019direct, gu2019stca}, among which the backprogation through time (BPTT) algorithm 
has become an effective way to train SNN models with high accuracy via global optimization. Table \ref{tab:STDP_BPTT} lists some reported accuracies for SNNs learnt by STDP and BPTT. Apparently, BPTT shows superior accuracy. For the more difficult datasets like CIFAR10, ImageNet, and CIFA10-DVS, STDP cannot provide good results while BPTT can do (see Table \ref{tab:STDP_BPTT} and Table \ref{tab:sparsity}).
Recent studies \cite{lillicrap2019backpropagation, lillicrap2020backpropagation} also try to reveal the connection between backpropagation and the brain, which is interesting but out of the scope of this work.


\textbf{Spike-based and Sparse Computing for High Efficiency.} BPTT is a costly learning algorithm with backpropagation across the entire network and all timesteps. 
Fortunately, we find opportunities after a detailed algorithm profiling.

\begin{table}[!htbp]
\caption{Sparsity in the backward pass of BPTT during SNN learning. Abbreviation: I-input, O-output.}
\label{tab:sparsity}
\vspace{5pt}
\centering
\renewcommand\arraystretch{1.3}
\resizebox{0.49\textwidth}{!}{
\begin{tabular}{c|c|c|c|c|c|c|c}
\hline \hline
\textbf{Dataset} & \textbf{Layer} & \textbf{conv1} & \textbf{conv2} & \textbf{conv3} & \textbf{conv4} & \textbf{conv5} & \textbf{conv6} \\
 
  \hline
MNIST & O sparsity (\%) & 22.66 & 85.99 & 52.74 & 74.74 & 53.82 & -\\
 
\cline{2-7}
Acc: 99.67\% & I sparsity (\%) & 99.18 & 97.46 & 98.76 & 97.44 & 97.93 & -\\

 \hline
CIFAR10 & O sparsity (\%) & 54.06 & 83.22 & 80.22 & 82.75 & 87.24 & -\\
 
\cline{2-7}
Acc: 87.63\% & I sparsity (\%) & 91.41 & 83.89 & 89.92 & 88.63 & 84.39 & -\\

 \hline
N-MNIST & O sparsity (\%) & 59.89 & 86.13 & 92.95 & - & - & -\\
 
\cline{2-5}
Acc: 98.97\% & I sparsity (\%) & 97.24 & 97.36 & 99.26 & - & - & -\\

 \hline
CIFAR10-DVS & O sparsity (\%) & 62.93 & 88.23 & 82.07 & - & - & -\\
 
\cline{2-5}
Acc: 63.00\% & I sparsity (\%) & 87.69 & 74.14 & 95.77 & - & - & -\\

 \hline
ImageNet & O sparsity (\%) & 17.97 & 16.21 & 14.65 & 19.82 & 16.65 & 15.09\\
 
\cline{2-8}
Acc: 60.90\% & I sparsity (\%) & 94.95 & 94.11 & 92.94 & 96.02 & 94.38 & 93.35\\

\hline \hline
\end{tabular}}
\end{table}

There are three phases in the BPTT learning: forward pass, backward pass, and weight update. In the forward pass and weight update, one of the two operands for the computational operations is a binary spike,
which implies that the costly multiplication units are no longer needed. In the backward pass, although multiplication operations cannot be avoided, there is rich sparsity that can be exploited. On one side, one of the operands for the computational operations in the backward pass is the potential gradient that is sparse; on the other side, the outputs are the gradients of spike activities, a part of which will be zeroed out when backpropagating through the firing function. The detailed BPTT for SNN learning can be found in Section \ref{sec:arch:BPTT}. In Table \ref{tab:sparsity}, we present the input and output sparsity values of each layer during backpropagation. We take several SNN models and datasets for illustration, and the detailed network configuration can be found in Table \ref{tab:net}. It can be seen that the sparsity is quite rich, indicating opportunities to reduce compute and storage.

%% file: text/architecture.tex
\section{Architecture Design}

\subsection{Analysis of BPTT for SNN Learning}\label{sec:arch:BPTT}
\label{accurate_learning}

In this subsection, we detail the BPTT learning process for SNNs \cite{wu2018spatio, wu2019direct, gu2019stca}. Table \ref{tab:variable} summarizes the commonly used variables. We use $\triangledown$ to denote the gradient of the loss function with respect to an intermediate variable, e.g., $\triangledown s=\frac{\partial L}{\partial s}$ and $\triangledown u=\frac{\partial L}{\partial u}$. 

\begin{table}[!htbp]
\caption{Definition of variables.}
\label{tab:variable}
\vspace{5pt}
\centering
\renewcommand\arraystretch{1.3}
\resizebox{0.425\textwidth}{!}{
\begin{tabular}{c|c||c|c}   
   \hline \hline
   \textbf{Var} & \textbf{Description} & \textbf{Var} & \textbf{Description}\\
  \hline
  $s$ & spike & $u$ & membrane potential\\
  \hline
  $\triangledown s$ & spike gradient & $\triangledown u$ & potential gradient\\
  \hline
$\triangledown \tilde{s}$ & spike gradient mask & 
$\triangledown \tilde{u}$ & potential gradient mask\\
  \hline
  $w$ & weight & $\triangledown w$ & weight gradient\\
  \hline \hline
\end{tabular}}
\vspace{-5pt}
\end{table}

\begin{table*}[!htbp]
\caption{I/O and the major operation of an SNN layer for each learning stage. The
variables in the binary format are marked in red.}
\label{tab:layer_operation}
\vspace{5pt}
\centering
\renewcommand\arraystretch{1.3}
\resizebox{0.98\textwidth}{!}{
\begin{tabular}{c|c|c|c|c}   
   \hline \hline
   \textbf{Stage} & \textbf{Inputs} & \textbf{Outputs} & Involved Equation & Major Operation \\
  \hline
  
  \textbf{Forward Pass} & $u_{t-1}^l$, \textcolor{red}{$s_{t-1}^{l}$}, \textcolor{red}{$s_{t}^{l-1}$}, $w^{l-1}$ & \textcolor{red}{$s_{t}^{l}$} , $u_{t}^{l}$ , \textcolor{red}{$\triangledown \tilde{s}_{t}^{l}$} & Eq. (\ref{equ:fp}) & $\textstyle\sum\nolimits_j \textcolor{red}{s_t^{l-1}[j]} w^{l-1}[j,i]$, spike-based Conv/MM \\
   \hline

  \textbf{Weight Update} & $\triangledown u_t^{l+1}$ , \textcolor{red}{$s_t^l$} & $\triangledown w^l$ & Eq. (\ref{equ:bp_w}) & $\textstyle\sum_t \triangledown u_t^{l+1} [j] \textcolor{red}{s_t^l [i]}$, spike-based Conv/MM\\ 
   \hline
   
  \textbf{Backward Pass} & $\triangledown u_{t+1}^{l}$ , $u_t^l$, $\triangledown u_{t}^{l+1}$, \textcolor{red}{$\triangledown \tilde{u}_t^{l+1}$}, $w^l$, \textcolor{red}{$\triangledown \tilde{s}_t^l$},  \textcolor{red}{$s_t^l$} & $\triangledown u_t^l$, \textcolor{red}{$\triangledown \tilde{u}_t^l$} & Eq. (\ref{equ:bp}) & $\textstyle\sum\nolimits_j \triangledown u_t^{l+1}[j] w^l[i,j]$, sparse FP16 Conv/MM \\

  \hline \hline
\end{tabular}}
\vspace{-10pt}
\end{table*}

\begin{figure}[!htbp]
\vspace{-5pt}
\centering
\includegraphics[width=0.38\textwidth]{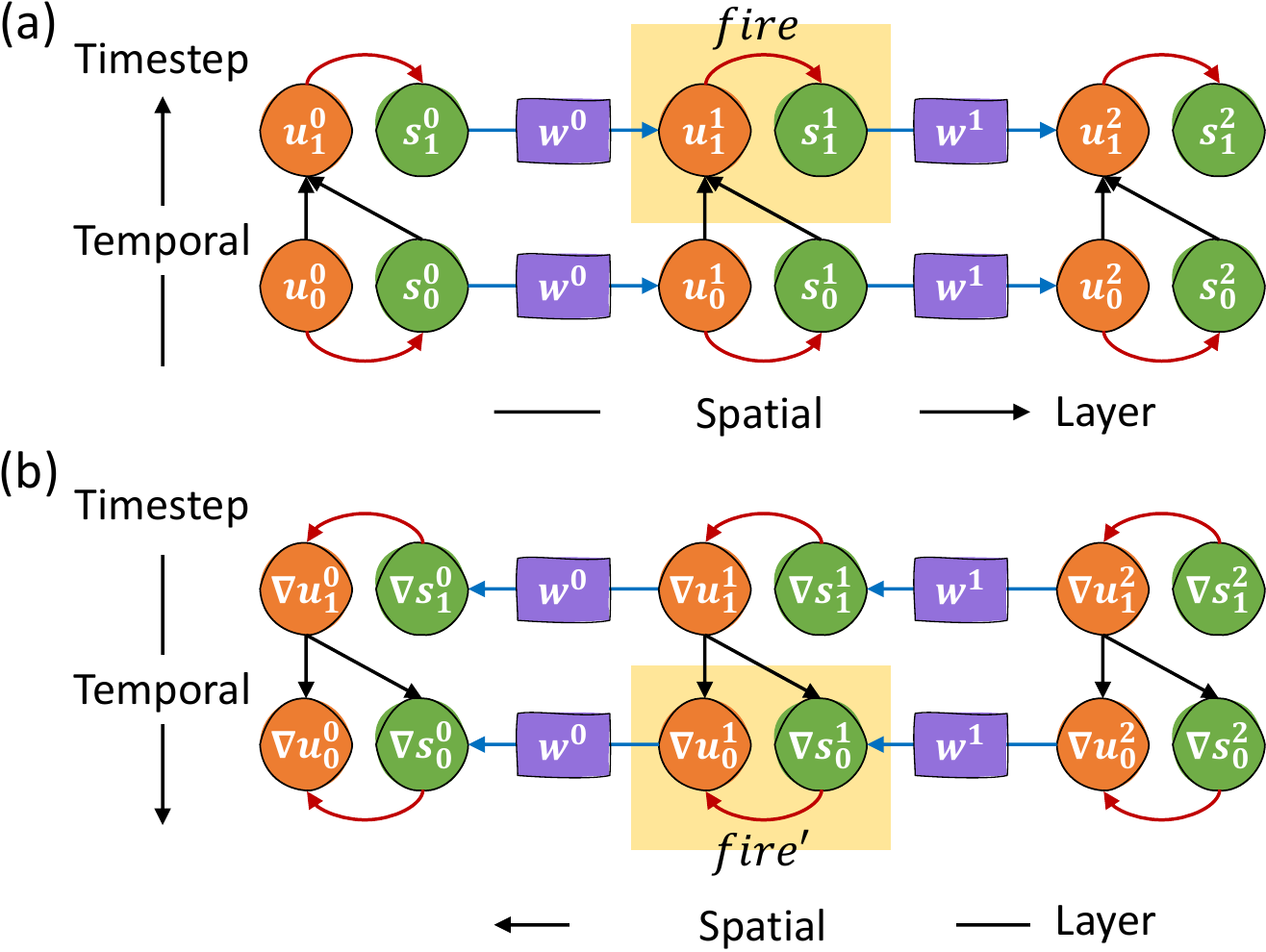}
\caption{Information propagation path of (a) the forward pass and (b) the backward pass in BPTT for SNN learning.}
\label{fig:BPTT}
\end{figure}

The neuronal behaviors of LIF-based SNNs \cite{gerstner2014neuronal} in the forward pass are governed by
\begin{equation}
\label{equ:fp}
\begin{cases}
    u_t^l[i]= 
    \underbrace
    {\alpha u_{t-1}^l[i] (1 - s_{t-1}^l[i])}_
    {temporal} + 
    \underbrace
    {\textstyle\sum\nolimits_j s_t^{l-1}[j] w^{l-1}[j,i]}_
    {spatial}, \\
    s_t^l[i]= fire(u_t^l[i]-th_f).
\end{cases}
\end{equation}
Here $u_t^l[i]$ and $s_t^l[i]$ represent the potential and spike of neuron $i$ in layer $l$ at timestep $t$, respectively. The potential update includes temporal and spatial parts. The temporal part is determined by the potential and spike in the previous timestep and a leakage factor $\alpha$; the spatial part is determined by the integration of weighted spikes from the previous layer. When the potential crosses a threshold $th_f$, the neuron fires a spike and resets its potential to zero. $fire(\cdot)$ is the Heaviside step function, i.e., $fire(x)=1$ if $x\geq0$; $fire(x)=0$ otherwise. Fig. \ref{fig:BPTT}(a) illustrates the propagation of the forward pass.

During the backward pass, the gradients of spike and potential can be expressed as
\begin{equation}
\label{equ:bp}
\begin{cases}
    \triangledown s_t^l[i]= 
    \underbrace
    {\triangledown u_{t+1}^l[i]
    (-\alpha u_t^l[i])}_{temporal} + 
    \underbrace
    {\textstyle\sum\nolimits_j \triangledown u_t^{l+1}[j] w^l[i,j] }_
    {spatial}, \\
    \triangledown u_t^l[i]= 
    \triangledown u_{t+1}^l[i] \alpha (1 - s_t^l[i]) + 
    \triangledown s_t^l[i] fire'(u_t^l[i]).
\end{cases}
\end{equation}
The spike gradient $\triangledown s_t^l[i]$ is also comprised of temporal and spatial parts. The temporal part is determined by the potential gradient and potential at the next and the current timestep, respectively. The spatial part is determined by the integration of weighted potential gradients from the next layer. The propagation in the  backward pass is illustrated in Fig. \ref{fig:BPTT}(b). The final weight gradient is calculated by 
\begin{equation}
    \label{equ:bp_w}
    \triangledown w^l[i,j] = \textstyle\sum_t \triangledown u_t^{l+1} [j] s_t^l [i].
\end{equation}

Originally, the derivative of the step function ($fire(\cdot)$) does not exist (a Dirac-like curve), which is known as the non-differentiability of the spike activity. In \cite{wu2018spatio}, a pulse curve is proposed to approximate the Dirac-like derivative, as follows
\begin{equation}
    \label{equ:fire}
    fire'(u_t^l[i]) \approx
    \begin{cases}
    \beta, & th_l < u_t^l[i] < th_r, \\
    0, & otherwise.
    \end{cases}
\end{equation}
This derivative implies that $\triangledown s_t^l[i]$ in the backward pass is valid only when $\triangledown u_t^l[i]$ in the forward pass lies in $[th_l, th_r]$.

We have the following observations for the BPTT learning: 

\begin{itemize} 
\item The spatial parts in Eq. (\ref{equ:fp}) \& (\ref{equ:bp}) and the weight gradient calculation in Eq. (\ref{equ:bp_w}) require the Conv or matrix multiplication (MM) operation in a Conv or FC layer, respectively. Other operations are element-wise, which have much fewer workloads. Therefore, our architecture design makes more efforts to accelerate the costly Conv or MM operations in the context of SNN learning.

\item The spikes are in the binary format, i.e. either 0 or 1. It is efficient to store the spike data in a compact format and use LUT-based operation to reduce computation.

\item Based on the $fire'(\cdot)$ in Eq. (\ref{equ:fire}), we can determine the valid neurons (marked by $\triangledown \tilde{s}$) that allow the gradient to pass through during the backward pass, according to their potential values in the forward pass. Specifically, when a neuron's potential is within $[th_l, th_r]$ in the forward pass, it is valid and needs to calculate its spike gradient in the backward pass; otherwise, we can skip the computation (termed as output sparsity in the Backward Engine design, see Section \ref{sec:bp}). During forward pass, there is also no need to store the potentials (for the use in the temporal part of Eq. (\ref{equ:bp})) of invalid neurons.

\item The goal of learning is to update weights of the model via calculating weight gradients. From Eq. (\ref{equ:bp_w}), we find that $\triangledown w$ only requires potential gradients and does not involve spike gradients. Therefore, $\triangledown s$ can be treated as intermediate data and merged into the $\triangledown u$ calculation.
\end{itemize}

We list inputs, outputs, and the major operation of an SNN layer for each training stage in Table \ref{tab:layer_operation}. The variables in the binary format are marked in red. We use bit masks $\triangledown \tilde{s}$ and $\triangledown \tilde{u}$ to indicate which neurons have valid spike gradients and non-zero potential gradients in the backward pass, respectively. 

\subsection{Architecture Design for Individual Engines}

In this subsection, we first introduce how to handle data with different formats and then detail the architecture design for each engine in \emph{H2Learn}. Unlike the training accelerators for ANNs that usually adopt one engine for all training stages \cite{qin2020sigma, tu2020evolver}, we design different engines for each SNN training stage. The philosophy behind this design is that the behavior of each training stage is distinct from each other. Specifically, (1) different data formats, i.e., one of the operands in the forward pass and weight update is a spike, however, all operands in the backward pass are real values; (2) different computation characteristics, i.e., the forward pass and weight update can take benefits from spikes to improve efficiency, however, the backward pass utilizes the input and output sparsity to simplify computation; (3) different dataflows, i.e., the Conv (or MM) dataflows in the forward and backward passes are distinct from that in weight update. Based on these special features, that are distinct from ANNs, we design a Forward Engine, a Weight Update Engine, and a Backward Engine, which form the backbone of our \emph{H2Learn}. Finally, these engines can be pipelined during training to gain optimized overall performance.

\vspace{-5pt}
\begin{figure}[!htbp]
\centering
\includegraphics[width=0.35\textwidth]{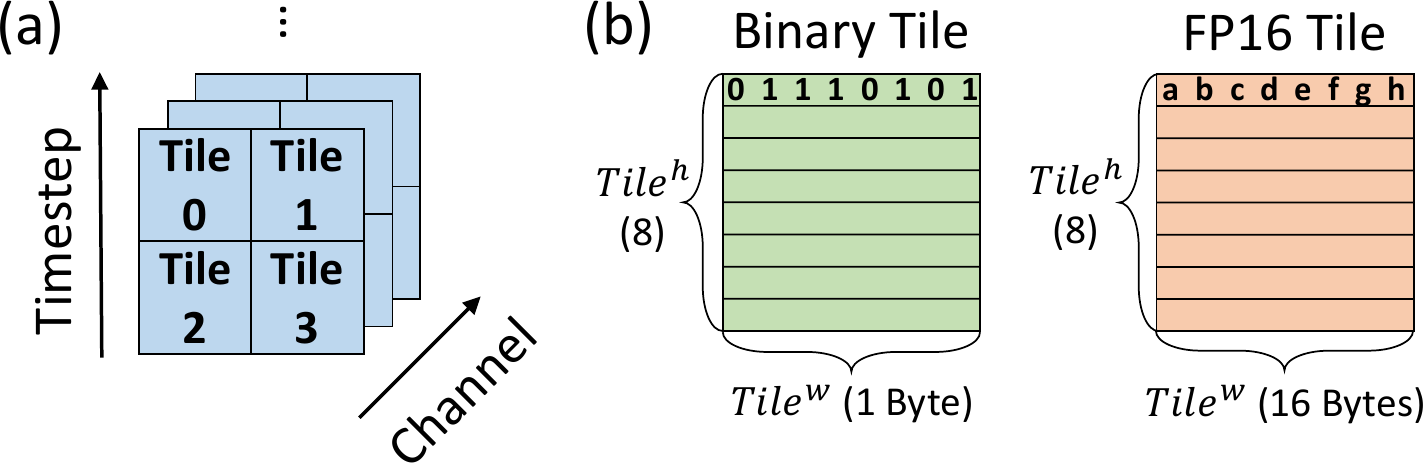}
\caption{Illustration of feature map (FM) tiling: (a) an example of FM tiling; (b) configuration of a tile with different data types.}
\label{fig:FM_tile}
\end{figure}

Fig. \ref{fig:FM_tile} shows the feature map (FM) tiling. The dimension of the FMs is $T \times C \times H \times W$, where $T$ represents the total number of timesteps, $C, H$ and $W$ stand for the channel, height, and width, respectively. For an FM locating at timestep $t$ and channel $c$, we split it into several tiles as shown in Fig. \ref{fig:FM_tile}(a), and the tile corresponds to the basic handling data unit in our design. In this work, we need to consider two data types: binary spike data and floating-point 16-bit (FP16) data. Fig. \ref{fig:FM_tile}(b) shows the example of a tile in the two data formats.

\begin{figure}[!htbp]
\vspace{-5pt}
\centering
\includegraphics[width=0.4\textwidth]{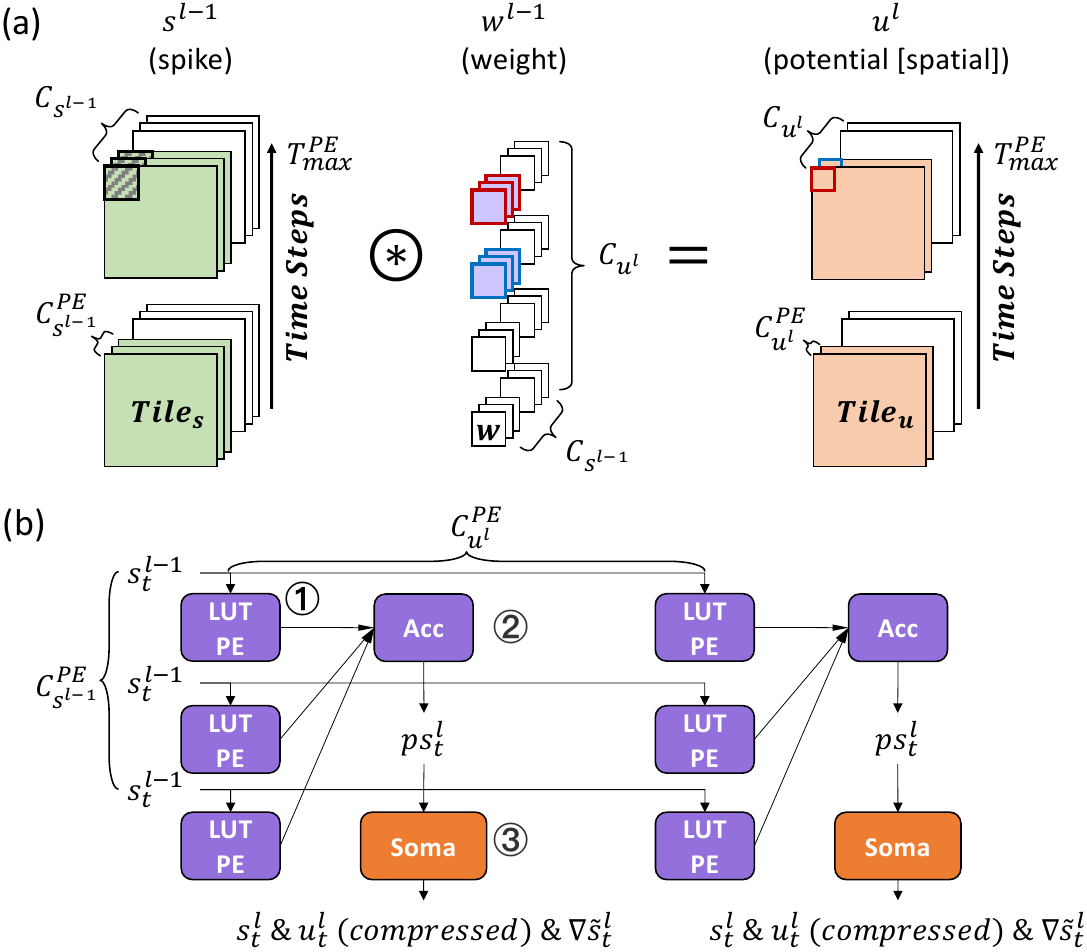}
\caption{Forward Engine: (a) operation; (b) microarchitecture.}
\label{fig:forward_engine1}
\end{figure}

\begin{figure*}[!htbp]
\centering
\includegraphics[width=0.925\textwidth]{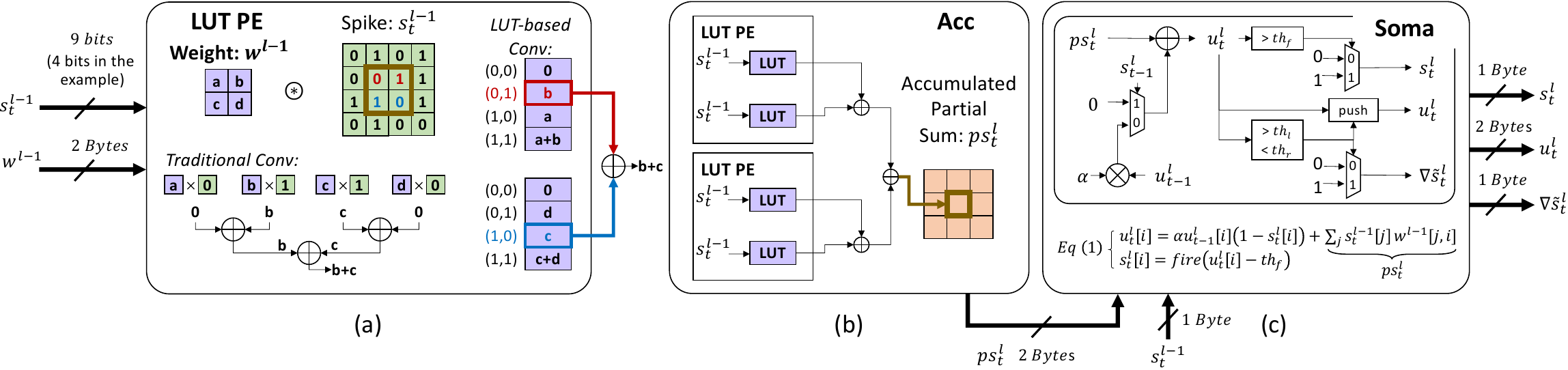}
\caption{Details of each function unit in Forward Engine: (a) LUT PE that performs a part of LUT-based Conv; (b) Acc unit that performs the rest of LUT-based Conv via inter-PE accumulation; (c) Soma unit that produces the spike, compressed potential, and spike gradient mask.}
\label{fig:forward_engine2}
\end{figure*}

\subsubsection{Forward Engine}

We design Forward Engine to handle the forward pass in BPTT learning. From Table \ref{tab:layer_operation}, the Conv of the spatial part in Eq. (\ref{equ:fp}) is the most costly operation, which is shown in Fig. \ref{fig:forward_engine1}(a). Each time, Forward Engine takes $T_{max}^{PE} \times C_{s^{l-1}}^{PE}$ tiles from $s_t^{l-1}$ and performs Conv with a part of weights whose size is $k^2 \times C_{s^{l-1}}^{PE} \times C_{u^{l}}^{PE}$ to produce $T_{max}^{PE} \times C_{u^{l}}^{PE}$ tiles of partial sums $ps_t^l$ which belong to the spatial part of $u_t^l$. $C_{s^{l-1}}$ and $C_{u^l}$ are the numbers of channels of $s^{l-1}$ and $u^l$, respectively. $T_{max}^{PE}$, $C_{s^{l-1}}^{PE}$, and $C_{u^l}^{PE}$ represent the maximum number of timesteps, channels of $s^{l-1}$ and $u^l$ that Forward Engine can process at one time. $k$ denotes the weight kernel size. In this work, we call such processing as one grid iteration.

Fig. \ref{fig:forward_engine1}(b) shows the microarchitecture of Forward Engine. In this example, the layout of the PE array is $C_{s^{l-1}}^{PE}$ rows by $C_{u^l}^{PE}$ columns. In Forward Engine, each row shares the same $Tile_s$ from $s^{l-1}$ and each column contributes to the same $Tile_u$ in $u^{l}$. The workflow of Forward Engine in a Conv layer includes following steps: \circled{1} the PE array receives sliding windows from $s_t^{l-1}$ and performs the LUT-based Conv (detailed later); \circled{2} each accumulator (Acc) integrates outputs from PEs of the same column; \circled{3} when the partial sum $ps_t^l$ includes all $C_{s^{l-1}}$ channels, the result will be sent to Soma to get $s_t^l$, $u_t^l$ (abandoned if out of $[th_l, th_r]$), and $\triangledown \tilde{s}_t^l$ (needed in the backward pass). The processing of different sliding windows, timesteps, and samples reuses the PE array resource.

Fig. \ref{fig:forward_engine2} details each block. Fig. \ref{fig:forward_engine2}(a) shows how to realize spike Conv using LUT. In this example, we perform a Conv between a $2 \times 2$ weight kernel and a sliding window in $s_t^{l-1}$. The 2D Conv is traditionally executed as a dot product. However, the inputs are binary spikes in SNNs, thus each sliding window can be represented as one of fixed states, i.e., $2^n$ states for $n$ elements. Here, 4 binary input elements in a sliding window have 16 patterns. We can calculate the Conv results for all possible patterns in advance and store them in an LUT. With this design, we use the input spikes as an access address to load results from the LUT. 

However, the LUT-based solution might increase the data need to store. We mitigate the storage consumption by splitting a large LUT into several small sub-LUTs. In Fig. \ref{fig:forward_engine2}(a), we use two sub-LUTs to cover different regions of the sliding window, and the LUT size in one PE can be reduced from 16 to 8. Notice that one extra adder is required to accumulate the partial results from sub-LUTs. We call this kind of PE units as LUT PE. Section \ref{sec:fe_wue} shows the detailed analyses for optimal LUT setting. In our design, each LUT PE loads the partial Conv results from all of its sub-LUTs, and the Acc unit accumulates the outputs from the LUT PEs in the same column using an adder tree with FP16 precision as Fig. \ref{fig:forward_engine2}(b). 

After finishing the spatial part compute, we feed the final $ps_t^l$ to the Soma unit which will update the potential and determine whether to fire a spike $s_t^l$ or not as in Fig. \ref{fig:forward_engine2}(c). According to our previous analysis, during the backward pass, we only need to store the potentials of valid neurons whose potentials fall into $[th_l, th_r]$. Therefore, Soma generates a compressed $u_t^l$ without storing zero elements and also produces the corresponding binary spike gradient mask $\triangledown \tilde{s}_t^l$:
\vspace{-5pt}
\begin{equation}
    \triangledown \tilde{s}_t^l[i] =
    \begin{cases}
    1, & th_l < u_t^l[i] < th_r, \\
    0, & otherwise.
    \end{cases}
\end{equation}

For FC layers, the weight matrix size is $C_{s^{l-1}} \times C_{u^l}$. Each column of the PE array belongs to a $u^l$ channel. We can treat the sub-LUTs in the same column as weight buffers that contain weights of the same output channel but many input channels. The number of input channels for each PE row relies on the size of sub-LUTs in each PE. During processing, each sub-LUT exports a weight element to Acc if the corresponding input spike is 1; otherwise exports 0. We do not compress $u_t^l$ in FC layers during the forward pass, since the data volume is far smaller than that in Conv layers. 

\subsubsection{Weight Update Engine}



\begin{figure}[!htbp]
\vspace{-5pt}
\centering
\includegraphics[width=0.385\textwidth]{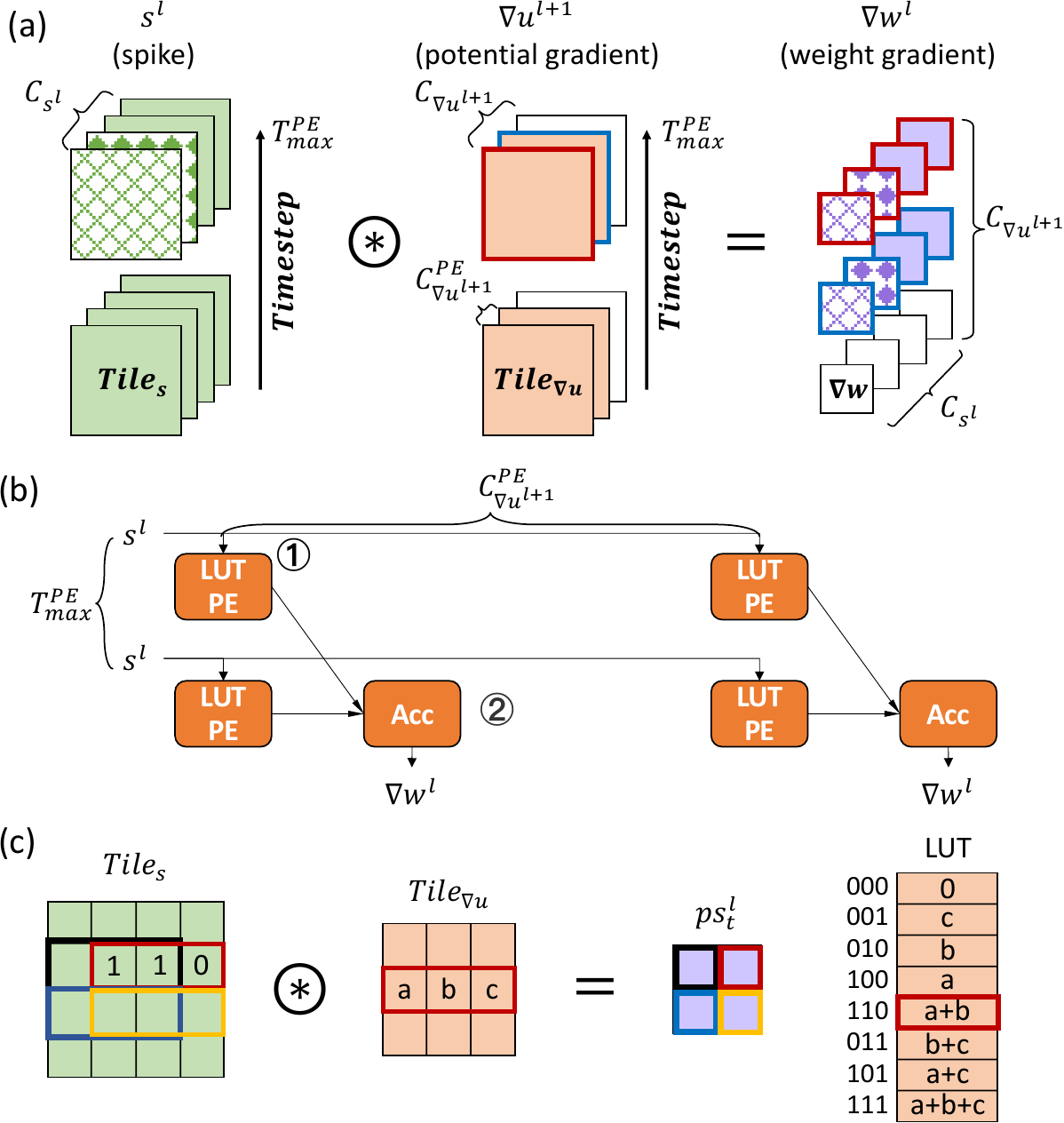}
\caption{Weight Update Engine: (a) operation; (b) microarchitecture; (c) LUT-based Conv.}
\label{fig:weight_update_engine}
\end{figure}

We design Weight Update Engine to calculate the weight gradient. From Eq.(\ref{equ:bp_w}), the weight gradient is calculated by performing a Conv between $s_t^l$ and $\triangledown u_t^{l+1}$, as shown in Fig. \ref{fig:weight_update_engine}(a). Weight Update Engine takes $T_{max}^{PE} \times C_{s^l}$ $Tile_s$ as inputs and performs Conv with $T_{max}^{PE} \times C_{\triangledown u^{l+1}}^{PE}$ $Tile_u$ to produce the partial sum $ps_t^l$ of $\triangledown w^l$ whose size is $k^2 \times C_{s^l} \times C_{\triangledown u^{l+1}}^{PE}$.

Fig. \ref{fig:weight_update_engine}(b) presents the microarchitecture of Weight Update Engine. The numbers of rows and columns correspond to $T_{max}^{PE}$ and  $C_{\triangledown u^{l+1}}^{PE}$, respectively.
The workflow of Weight Update Engine includes two steps: \circled{1} performs LUT-based Conv between $s^l$ and $\triangledown u^{l+1}$ in the PE array; \circled{2} accumulates the outputs from LUT PEs of the same column in the Acc unit. Notice that the processing of different sliding windows and input channels reuses the PE array resource. Since $s^l$ is in the binary format, we can still use the LUT-based Conv as in the Forward Engine. Fig. \ref{fig:weight_update_engine}(c) shows an example of LUT-based Conv in the Weight Update Engine.

For FC layers, the weight gradient calculation requires a dot product between two matrices whose sizes are $C_{s^l} \times T_{max}^{PE}$ and $T_{max}^{PE} \times C_{\triangledown u^{l+1}}$. We directly feed elements in $\triangledown u^{l+1}$ to the sub-LUTs. Every PE row shares inputs at the same timestep. Similar to Forward Engine, each sub-LUT exports an element to the Acc unit based on the state of the spike input.

\begin{figure}[!htbp]
\vspace{-5pt}
\centering
\includegraphics[width=0.425\textwidth]{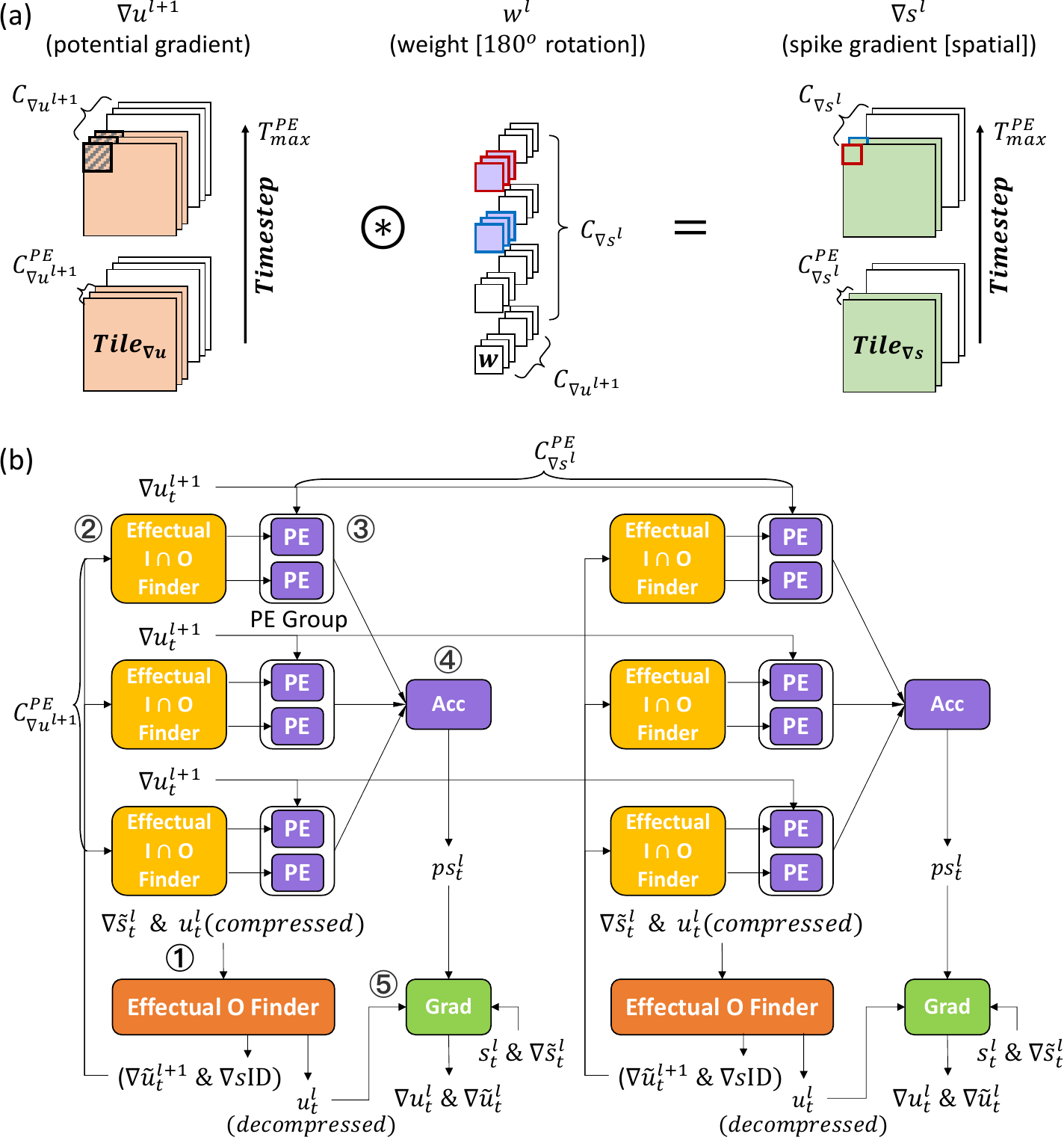}
\caption{Backward Engine: (a) operation; (b) microarchitecture.}
\label{fig:backward_engine1}
\end{figure}

\begin{figure*}[!htbp]
\vspace{-5pt}
\centering
\includegraphics[width=0.9\textwidth]{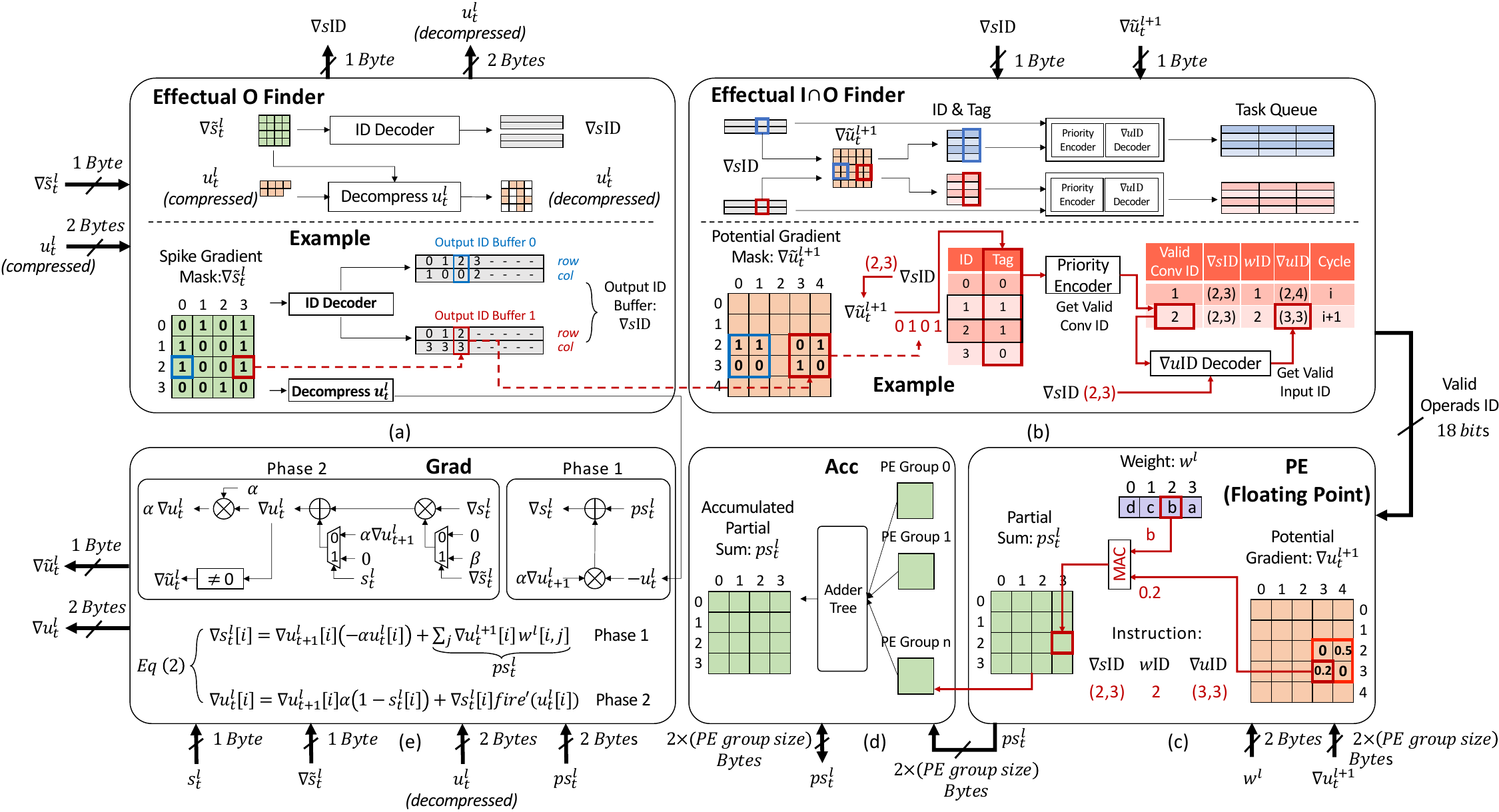}
\caption{Details of each function unit in Backward Engine: (a) Effectual Output Finder unit finds neuron IDs that have valid $\triangledown s$ and decompresses $u_t^l$, exploiting output sparsity; (b) Effectual Input$\cap$Output Finder unit generates valid microinstructions for each Conv operation in the PE unit, exploiting both input and output sparsity; (c) \& (d) PE and Acc units that perform the valid FP16 MACs in the Conv operation; (e) Grad unit that produces $\triangledown u_t^l$ and $\triangledown \tilde{u}_t^l$.}
\label{fig:backward_engine2}
\end{figure*}

\subsubsection{Backward Engine}\label{sec:bp}
We design Backward Engine to compute the potential gradient in the backward pass. From table \ref{tab:layer_operation}, Conv is the major operation, as shown in Fig. \ref{fig:backward_engine1}(a). Backward Engine takes $T_{max}^{PE} \times C_{\triangledown u^{l+1}}^{PE}$ $Tile_{\triangledown u}$ and performs Conv with corresponding weight kernels (after $\ang{180}$ rotation) whose size is $k^2 \times C_{\triangledown u^{l+1}}^{PE} \times C_{s^l}^{PE}$ to generate $T_{max}^{PE} \times C_{s^l}^{PE}$ tiles of partial sum $ps_t^l$ for $\triangledown s_t^l$. The operand data type here is FP16, which increases the compute cost. However, we can exploit the output sparsity (indicated by  $\triangledown \tilde{s}_t^l$ pre-generated in the Soma unit during the forward pass). Moreover, we can utilize the input sparsity in $\triangledown u$ to further simplify computation. Specifically, we use a bitmap $\triangledown \tilde{u}$ to record the input sparsity information:
\begin{equation}
    \triangledown \tilde{u}_t^l[i] =
    \begin{cases}
    1, & \triangledown u_t^l[i] \neq 0, \\
    0, & otherwise.
    \end{cases}
\end{equation}
From the hardware perspective, we use input and output neuron IDs in a tile to locate valid operands according to the corresponding bitmaps, which will be introduced latter.

The microarchitecture of Backward Engine is shown in Fig. \ref{fig:backward_engine1}(b). The PE array layout is $C_{\triangledown u^{l+1}}^{PE}$ rows by $C_{s^l}^{PE}$ columns. PEs in the same row share the same $Tile_{\triangledown u}$ from $u^{l+1}$ and the corresponding $\triangledown \tilde{u}_t^{l+1}$. PEs in the same column generate the partial sum $ps_t^l$ of the spatial part of the same $Tile_{\triangledown s}$ of $s^{l}$, also these PEs share the same $\triangledown \tilde{s}_t^l$. The workflow of Backward Engine in a Conv layer has the following steps: \circled{1} Effectual O Finder gets the valid output neuron IDs ($\triangledown s$IDs) for Conv according to $\triangledown \tilde{s}_t^l$, and decompresses the compressed $u_t^l$ into the original dense format; \circled{2} Effectual I$\cap$O Finder further generates valid input and output neuron IDs for Conv according to $\triangledown \tilde{u}_t^{l+1}$ and the above $\triangledown s$IDs; \circled{3} PEs perform Conv that exploits both input and output sparsity; \circled{4} the Acc unit accumulates partial sums from PEs of the same column; \circled{5} the Grad unit calculates $\triangledown s_t^l$ and $\triangledown u_t^l$. Finally, the produced $\triangledown u_t^l$ and the corresponding $\triangledown \tilde{u}_t^l$ serve as outputs. The processing of different sliding windows, timesteps ($T_{max}^{PE}$), and samples reuses the PE array resource.

Fig. \ref{fig:backward_engine2} shows each function unit in Backward Engine. 
We consider multi-PEs 
in a PE group to improve the parallelism of processing a tile. The first functionality of the Effectual O Finder unit is to get the valid output neuron IDs ($\triangledown s$IDs) according to the stored spike gradient mask $\triangledown \tilde{s}_t^l$ in the forward pass. As an example shwon in Fig. \ref{fig:backward_engine2}(a), we take two Output ID Buffers to store the coordinates of $\triangledown s$IDs, which are shared by the PE groups in the same column. ID Decoder scans the elements in $\triangledown \tilde{s}_t^l$ row by row, and then writes $\triangledown s$IDs into Output ID Buffers alternatively. In this way, the two buffers can save close amount of $\triangledown s$IDs that need to be processed by PEs in a PE group. Each PE in a PE group would process the workloads stored in one of the buffers.
The second functionality of Effectual O Finder is to decompress $u_t^l$ that is stored in a compressed form during the forward pass. This decompression can make the element-wise operations in the Grad unit easier to execute.

Then, each Effectual O Finder unit sends $\triangledown s$IDs to the Effectual I$\cap$O Finder units in the same column to search valid multiplications in Conv (termed as valid Conv IDs) wherein both input ($\triangledown \tilde{u}_t^{l+1}$) and output ($\triangledown \tilde{s}_t^l$) are valid (i.e., non-zero). The procedure is shown in Fig. \ref{fig:backward_engine2}(b). We process two Output ID Buffers separately (marked in blue and red). For example, computing the point $\triangledown s$ID=(2,3) (marked in red) needs a Conv between the sliding window $\triangledown {u}_t^{l+1}[2:3, 3:4]$ and the weight kernel. Given the binary potential gradient mask ($\triangledown \tilde{u}_t^{l+1}$), we use a Tag to indicate the state (valid/invalid) of the elements in the sliding window. Next, the Priority Encoder produces a valid Conv ID per cycle based on the Tag. The valid weight value can be found through $w$ID which corresponds to the valid Conv ID and $\triangledown u$ID can be easily acquired based on the Conv ID and $\triangledown s$ID in $\triangledown u$ID Decoder.

Next, PE units perform the valid MACs. Fig. \ref{fig:backward_engine2}(c) shows an example of the workflow in one PE. Each PE in a PE group executes the MACs according to one of the tasks in the corresponding Effectual I$\cap$O Finder. The weight from $w^l$ and the input from $\triangledown {u}_t^{l+1}$ are read according to $w$ID and $\triangledown u$ID, respectively; the MAC result is written into the partial sum ($ps_t^l$) buffer. In real implementation, PEs in the same PE group write the result to independent partial sum buffers. The accumulated results in a tile will be reordered in the Acc unit. 
Different PEs in the same PE group reuse the weight buffer. 
After all PE groups complete the Conv of a tile, PE groups in the same column send their calculated partial sums to the Acc unit for accumulation. Note that, 
PEs in the PE array work asynchronously during execution but synchronize after all PE groups complete the computation of a tile.

At last, we calculate $\triangledown s_t^l$ and $\triangledown u_t^l$ in the Grad unit. As in Eq. (\ref{equ:bp}), we split the gradient calculation into two phases. In phase 1, Grad takes $ps_t^l$, $u_t^l$ and $\triangledown u_t^{l+1}$ to generate $\triangledown s_t^l$. In phase 2, besides the calculation of $\triangledown u_t^l$, we need to generate the potential gradient mask $\triangledown \tilde{u}_t^l$ that reflects the input sparsity of the next backpropagated layer. These two phases can be easily implemented with element-wise operations.

For FC layers, we do not consider any sparsity, since the computation workloads in FC layers are much fewer than those in Conv layers. We disable Effectual O Finders and Effectual I$\cap$O Finders when performing FC layers. In the PE array, the weight buffers in the same column store the weights of different $\triangledown u_t^{l+1}$ channels but of the same $\triangledown s_t^{l}$ channel.

\subsection{Overall Architecture and Pipeline}

\subsubsection{Overall Architecture}

\begin{figure}[!htbp]
\vspace{-5pt}
\centering
\includegraphics[width=0.485\textwidth]{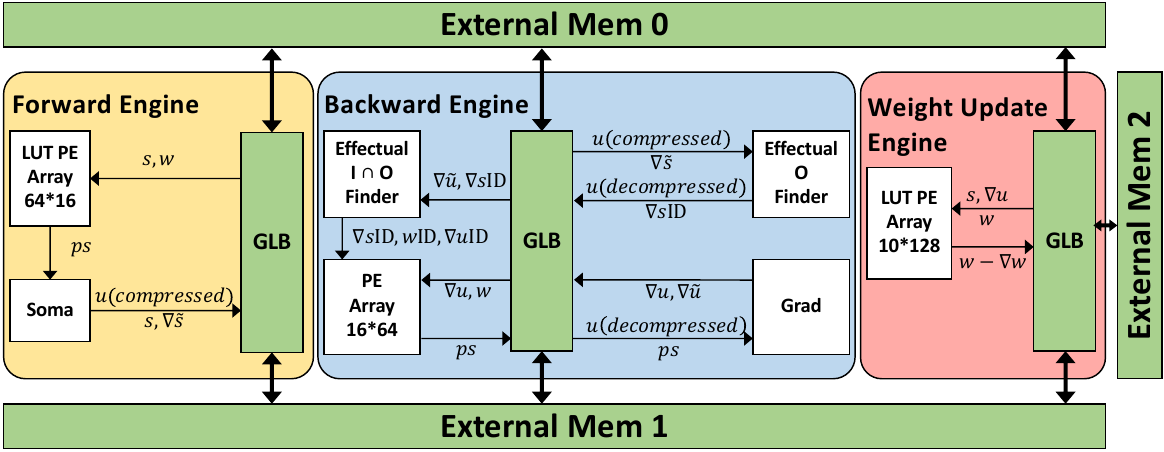}
\vspace{-15pt}
\caption{Overall architecture of \emph{H2Learn}.}
\label{fig:overall_arch}
\end{figure}

Fig. \ref{fig:overall_arch} shows the overall architecture of \emph{H2Learn}. In order to reduce the conflicts in data load and store, we use three external memory spaces for simplicity. Specifically, Mem 0 and Mem 1 are used to store spikes, potentials, potential gradients, gradient masks, and weights. During the current forward pass, Forward Engine writes the results into Mem 0(1); while in the next forward pass, the results will be alternatively written into Mem 1(0). Weight Update Engine and Backward Engine request the saved data in the forward pass as their inputs. Mem 2 is designed to store weight gradients that are only used by Weight Update Engine. In real implementation, a single external memory space with a high-bandwidth arbiter is also a possible solution.

We consider the scalability of each processing engine in two directions. The first direction is to increase the size of PE arrays. However, with this change, we should also increase the capacity of global buffers and the off-chip memory bandwidth. Notice that the numbers of rows and columns in each PE array correspond to the number of FMs and timesteps in Conv layers, thus a too large PE array size will decrease the resource utilization for a given Conv layer. Another direction is to increase the number of compute units in a PE group and the number of Acc units. Specifically, in Forward Engine and Weight Update Engine, we increase the number of Acc units to process multiple tiles simultaneously; in Backward Engine, we increase both the number of compute units in a PE group and the number of Acc units. Besides the above scalability directions, we can further use multiple \emph{H2Learn}s to build a distributed system. Each of them runs the learning algorithm with different input samples, and the weight gradients are gathered after all of them finish training. 

\subsubsection{Execution Pipeline}

\begin{figure}[!htbp]
\vspace{-5pt}
\centering
\includegraphics[width=0.485\textwidth]{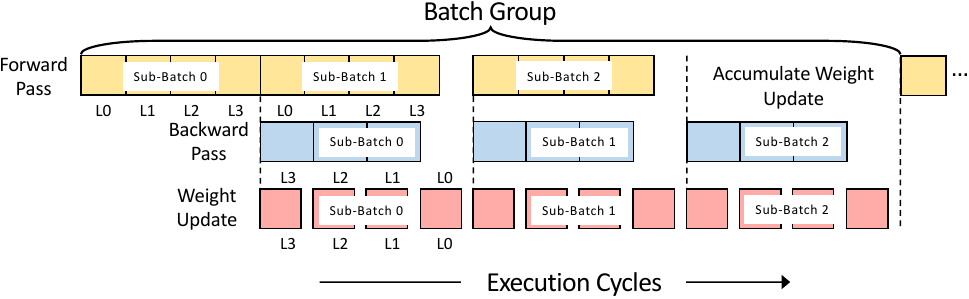}
\caption{Overall pipeline during learning.}
\label{fig:pipeline}
\end{figure}

Fig. \ref{fig:pipeline} shows the execution flow of training. The yellow, blue, and red boxes indicate the execution in Forward Engine, Backward Engine, and Weight Update Engine, respectively. The sub-batch represents the batch size that each engine can process at a time. The batch group represents the number of sub-batches for the update of weights. Although every sub-batch involves the calculation of weight gradients, only the last sub-batch in a batch group triggers the weight update. Therefore, the overall batch size can be flexibly reconfigured by adjusting the batch group size. During training, the three engines can be pipelined for higher throughput, i.e., when Backward Engine and Weight Update Engine are processing the current sub-batch, Forward Engine can process the next sub-batch. To enhance the overlap between forward and backward passes and reduce the amount of data need to store in the forward pass, each sub-batch should contain as few samples as possible. We set the sub-batch size to 4 in our design to well utilize the hardware resource. 

%% file: text/result.tex
\section{Evaluation}

\subsection{Experimental Setup}
Our experiments focus on pattern recognition tasks in both image and spike based datasets that are widely used for SNN evaluation. The image-based datasets include MNIST \cite{lecun1998gradient}, CIFAR10 \cite{krizhevsky2009learning} and ImageNet \cite{deng2009imagenet} that are sampled to spikes; while the spike-based datasets include N-MNIST \cite{orchard2015converting} and CIFAR10-DVS \cite{li2017cifar10} that are originally acquired through DVS \cite{Lichtsteiner2008A}. The network configurations are detailed in Table \ref{tab:net}.  

\begin{table}[!htbp]
\caption{Network configurations. $T$ is set to 10.}
\label{tab:net}
\vspace{5pt}
\centering
\renewcommand\arraystretch{1.3}
\resizebox{0.47\textwidth}{!}{
\begin{tabular}{c|c|c}
\hline \hline
 \textbf{Dataset} & \textbf{Input Size} & \textbf{Network Structure} \\
 
 \hline
N-MNIST & $32*32*2*T$ &
128C3-128C3-AP2-384C3-384C3-AP2-\\
\cline{1-2}
CIFAR10-DVS & $42*42*2*T$ & 512FC-512FC-10FC\\
 
 \hline
MNIST & $28*28*1*T$ &
64C3(Encoding)-128C3-AP2-256C3-256C3-AP2-\\
\cline{1-2}
CIFAR10 & $32*32*3*T$ & 512C3-512C3-512FC-512FC-10FC\\

\hline
\multirow{2}*{ImageNet} & \multirow{2}*{$224*224*T$} & 64C3S2(Encoding)-128C3S2-256C3S2-256C3S2-\\
& & 384C3-256C3-256C3S2-4096FC-4096FC-1000FC\\

\hline \hline
\end{tabular}}
\end{table}

We build cycle accurate simulators for \emph{H2Learn} and the accelerator baselines in our experiments. The area and energy are measured through synthesized implementations. We implement \emph{H2Learn}'s RTL and synthesize it in Synopsis Design Compiler with TSMC 28nm library. The area and energy of GLBs are estimated via Cacti \cite{li2011cacti}. In our simulation, we evaluate the average energy per operation for all compute and storage units, and the total energy is obtained by estimating the number of operations.

\begin{table}[!htbp]
\caption{Specifications of engines in \emph{H2Learn}.}
\label{tab:specification}
\vspace{5pt}
\renewcommand\arraystretch{1.3}
\resizebox{0.475\textwidth}{!}{
\begin{tabular}{c|c|c}
\hline \hline

& LUT PE Array Size & 64 (rows) $\times$ 16 (cols) \\
& LUT PE & 3 Sub-LUTs/PE, 16 Bytes/Sub-LUT \\
\textbf{Forward} & Acc & 3$\times$64$\times$16 (3072) adders, parallism=4 \\
\textbf{Engine} & \# Somas & 16 \\
& GLB & 503 KB \\
& Area & 21.61 mm$^2$ \\
\hline

& LUT PE Array Size & 10 (rows) $\times$ 128  (cols) \\
& LUT PE & 2 Sub-LUTs/PE, 32 Bytes/Sub-LUT \\
\textbf{Weight Updte}& Acc & 2$\times$10$\times$128 (2560) adders, parallism=4 \\
\textbf{Engine} & GLB & 2684 KB \\
& Area & 22.68 mm$^2$ \\
\hline

& PE Group Array Size & 16 (rows) $\times$ 64  (cols) \\
& PE Group Size & 4 \\
& \# Effectual O Finders & 64 \\
\textbf{Backward} & \# Effectual I$\cap$O Finders & 16$\times$64$\times$4\\
\textbf{Engine} & PE & 16$\times$64$\times$4 MAC\\
& Acc & 16$\times$64$\times$4 adders\\
& \# Grads & 64 \\
& GLB & 4840.5 KB \\
& Area & 66.17 mm$^2$ \\

\hline \hline
\end{tabular}}
\end{table}

The configuration of all engines in \emph{H2Learn} are listed in Table\ref{tab:specification}. In our evaluation, we build a different baseline model for each engine. Specifically, for Forward Engine and Weight Update Engine, the baseline models adopt non-LUT implementation which consume 36,864 and 40,960 adders, under the parallelism of 4. For the baseline model of Backward Engine, the Effectual O Finder and Effectual I$\cap$O Finder units are removed and we do not exploit any sparsity.


Since \emph{H2Learn} focuses on the training scenario of SNNs, our final target comparison platform is the modern GPU, which is the backbone hardware for training. Table \ref{tab:high_level} compares the overall specifications between \emph{H2Learn} and NVIDIA V100 GPU \cite{nvidia2017v100}. We will demonstrate that \emph{H2Learn} can achieve substantial speedup and energy efficiency improvement with far less area consumption in Section \ref{sec:compare_h2learn}. 

\begin{table}[!htbp]
\caption{Specifications of \emph{H2Learn} and NVIDIA V100 GPU.}
\label{tab:high_level}
\vspace{5pt}
\centering
\renewcommand\arraystretch{1.3}
\resizebox{0.4\textwidth}{!}{
\begin{tabular}{c|c|c}
\hline \hline
&\textbf{\emph{H2Learn}} & \textbf{\emph{GPU V100}} \\
\hline
Technology & TSMC 28 nm & TSMC 12 nm\\
Area & 110.46 mm$^2$ & 815 mm$^2$\\
Clock Frequency & 800 MHz & 1530 MHZ\\
Off-chip Memory Bandwidth & 128 GB/s$\times$3 & 900 GB/s\\
Throughput & 27.85 TFLOPS & 15.7 TFLOPS\\
Power & 20.57 W & 300 W \\
\hline \hline

\end{tabular}}
\end{table}

\subsection{Evaluation of Engines in H2Learn}
\label{sec:engine_analyse}

\subsubsection{Forward Engine \& Weight Update Engine}\label{sec:fe_wue}

\begin{figure}[!htbp]
\centering
\includegraphics[width=0.485\textwidth]{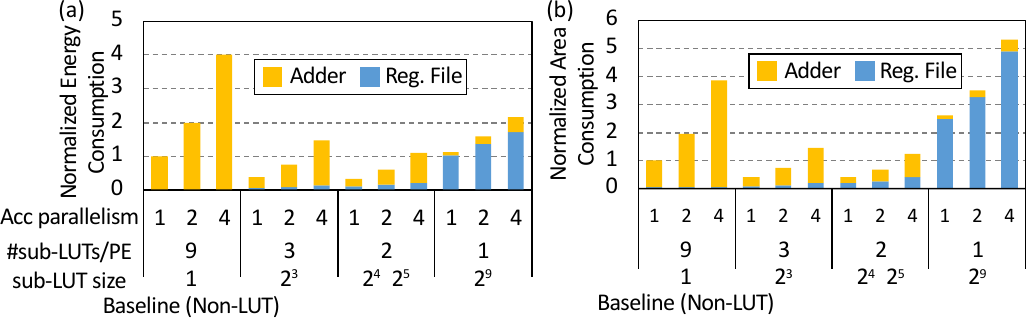}
\caption{Evaluation of LUT PE in Forward Engine under different configurations: (a) area and (b) energy.}
\label{fig:result_lut}
\end{figure}

Now, we evaluate the LUT-based Engines, i.e., Forward Engine and Weight Update Engine. Fig. \ref{fig:result_lut} shows how the PE configuration affects the area and energy consumption, where both PEs and Acc units are considered. The baseline is a non-LUT design. The LUT configuration is determined by the number of sub-LUTs per PE (\#sub-LUTs/PE) and the size of each sub-LUT (sub-LUT size). We assume that the size of each sliding window in Conv is 3$\times$3. We find that when we decrease the \#sub-LUTs/PE, the area and energy of computation units (i.e., adders) are reduced, however, the requirement for register files is increased. This is actually a trade-off, i.e., fewer sub-LUTs per PE can save more adders for compute but require more register files for storage.

We also estimate the area and energy consumption with different parallelism settings in Acc units. Here the parallelism means that multiple tiles are simultaneously processed in the LUT PE array and the resources for Acc units are copied accordingly. From the results, it can be seen, besides the increased adders in Acc units, the requirement for register files is also increased with a slower slope, since we do not scale up the number of sub-LUTs but use MUX to make sure multiple elements can be read from each sub-LUT in the meantime. 

In our design, considering the unified sub-LUT size and lower overhead, we set the \#sub-LUTs/PE to 3 and the sub-LUT size to 8 in Forward Engine, corresponding to 3$\times$3 sliding windows. In Weigh Update Engine, the \#sub-LUTs/PE is 2 and the sub-LUT size is 16, corresponding to 1$\times$8 sliding windows. For a larger Conv kernel size, we can partition it into multiple smaller kernels and map onto multiple LUT PEs. The overall LUT sizes in Forward Engine and Weight Update Engine are 48 KB and 80 KB, respectively.

\begin{figure}[!htbp]
\centering
\includegraphics[width=0.485\textwidth]{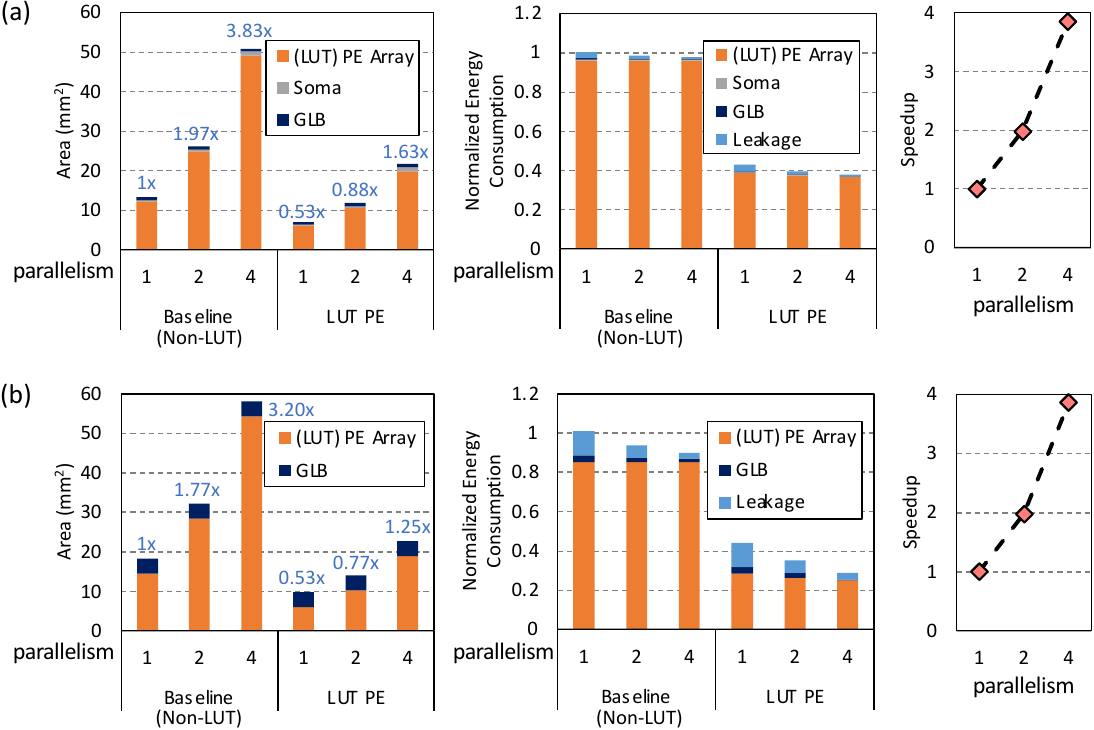}
\caption{Area, energy, and throughput of (a) Forward Engine and (b) Weight Update Engine.} 
\label{fig:result_lut_engine}
\end{figure}

Next, we analyze the entire Forward Engine and Weight Update Engine when performing a Conv layer. The size of $s$, $u$, and $\triangledown u$ is $4\times10\times256\times56\times56$ ($N \times T \times C_s (\text{or}~C_u,~\text{or}~C_{\triangledown u}) \times W \times H$), and the size of $w$ is $3\times3\times256\times256$ ($k \times k \times C_s \times C_u(\text{or}~C_{\triangledown u})$). The results are shown in Fig. \ref{fig:result_lut_engine}, where the baseline architecture adopts the naive accumulation-based rather than LUT-based PE. We have the following observations: (1) The PE array consumes most of the area and the LUT-based design can significantly reduce the area overhead; (2) Although the outputs of both Forward Engine and Weight Update Engine are in FP16, the number of columns in the PE array of Weight Update Engine is much larger, leading to a larger GLB size; (3) The energy consumption with different parallelism setting is close, because the amount of total workloads under different parallelism is identical; (4) The throughput can be improved as the parallelism increases and the leakage energy can be reduced; (5) Compared to the baseline architecture when the parallelism equals 4, our LUT-based solution can achieve 2.35$\times$ area saving, 2.58$\times$ energy saving in Forward Engine and 2.56$\times$ area saving, 3.00$\times$ energy saving in Weight Update Engine.

\subsubsection{Backward Engine}

\begin{figure}[!htbp]
\vspace{-5pt}
\centering
\includegraphics[width=0.47\textwidth]{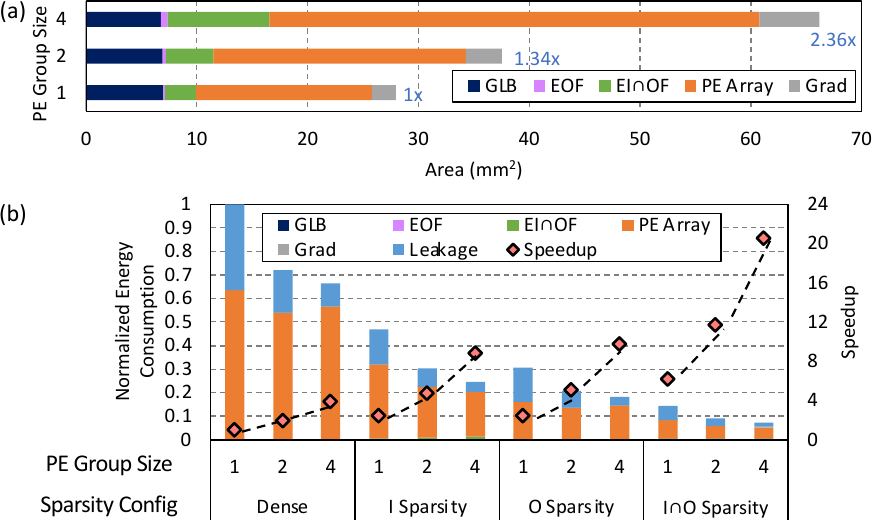}
\caption{Backward Engine evaluation: (a) area overhead; (b) energy consumption and speedup.}
\label{fig:result_backward_engine}
\end{figure}

We adopt the same Conv layer as we used in Section \ref{sec:fe_wue}. As depicted in Fig. \ref{fig:result_backward_engine}(a), as the PE group size grows, the area overhead increases but the GLB size does not change obviously since the resulting output volume keeps the same. In Fig. \ref{fig:result_backward_engine}(b), we measure the energy and throughput, where we set both the sparsity of $\triangledown \Tilde{s}^{l}$ (output sparsity) and $\triangledown \Tilde{u}^{l+1}$ (input sparsity) to 75\%. We first build a dense baseline model without considering any input and output sparsity. We also adopt our architecture to exploit only input or output sparsity as two other baselines. From the results, we find that the leakage consumes a huge amount of energy, which mainly comes from GLB. Another observation is that the energy consumption of the PE array (including the Acc units) occupies the most in the dense architecture, because it cannot bypass any computation. Also, the energy consumed by PE array is much higher when we consider the input sparsity only, since more accumulations of the partial sums are needed when compared with those considering the output sparsity. Since the sparsity settings of $\triangledown \Tilde{s}^{l}$ and $\triangledown \Tilde{u}^{l+1}$ are the same, the speedup results are similar when we consider the input or output sparsity only. Finally, when we consider both the input and output sparsity
, we can achieve 5.19$\times$ speedup and 9.24$\times$ energy saving compared with the dense baseline architecture.

\subsubsection{Design Space Analysis}

Table \ref{tab:specification} shows the configurations of \emph{H2Learn}. We adopt output stationary dataflow in all engines. Since the inputs of Forward Engine are binary spikes that are more compact than the outputs, we set a larger number of rows (64) in the PE array. Because of the output stationary dataflow, the result is written to external memory for every $C_s / 64$ grid iterations, which can help reduce the data traffic. Note that we use ping-pong buffer in GLBs. In Backward Engine, both inputs and outputs are in the FP16 format. We shrink the number of rows (16) but increase the number of columns (64), such that the amount of inputs needing to feed is reduced and the outputs can take longer time ($C_{\triangledown u}$/16 grid iterations) to be written to external memory. For Weight Update Engine, the number of rows is set to 10 (allowing to deal with 10 timesteps during a grid iteration), and the number of columns (128) is selected according to the PE array sizes in other two engines.

\begin{figure}[!htbp]
\vspace{-5pt}
\centering
\includegraphics[width=0.4\textwidth]{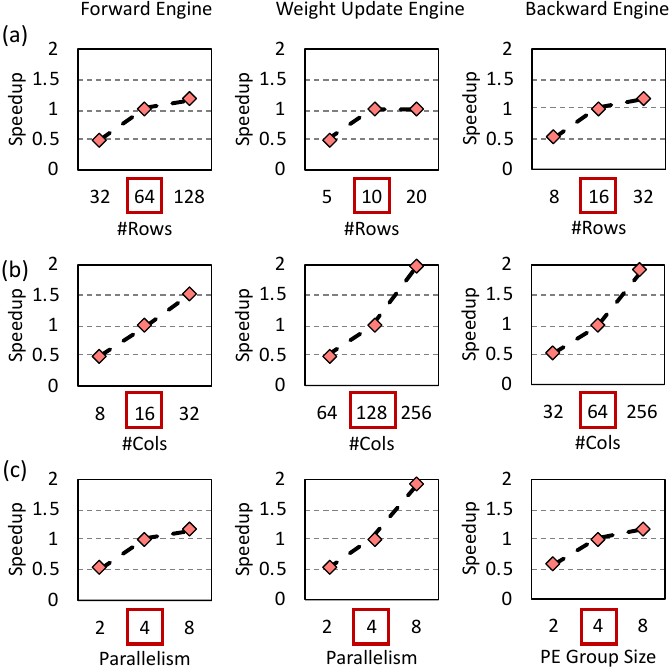}
\caption{Evaluation of engines in \emph{H2Learn} with different (a) number of PE array rows, (b) number of PE array columns, and (c) parallelism.}
\label{fig:result_design_space}
\end{figure}

Fig. \ref{fig:result_design_space} estimates the throughput of each engine in \emph{H2Learn} with different architecture configurations, including the number of PE array rows, columns, and the parallelism. The configurations within the red boxes are our optimal selections for each engine in Table \ref{tab:specification}, which consider both optimal performance and balanced compute resources in different engines. The optimal selection can fully utilize but will not be blocked by the off-chip memory bandwidth. Notice that when we evaluate one architecture configuration, we will fix the other two at the optimal settings. 
We also adopt the same Conv layer for evaluation as in Section \ref{sec:fe_wue}. From the results, we find that the optimal settings can always gain a 2$\times$ speedup when compared with the corresponding halved settings. This implies that the external memory bandwidth can satisfy our optimal settings. However, if we keep scaling up the PE array size or the parallelism, the throughput gain would degrade. In Forward Engine, all architecture configurations cannot get another 2$\times$ speedup by doubling the optimal settings. In Weight Update Engine, each PE array row deals with a unique timestep, thus the increase of extra rows is useless if the number of timesteps is small; however, the throughput can be doubled when we scale up the PE array columns or the parallelism. The reason is that, the final outputs of Weight Update Engine is the weight gradients which have a small volume and can stay on-chip until the entire Conv across all FMs finished. This lowers the bandwidth requirement and leaves room for the increase of compute resources. For Backward Engine, the inputs (i.e., $\triangledown u$) become the key factor to determine the external memory bandwidth requirement, because the input data type is FP16 and the load of inputs across all channels is more frequently than the write of stationary outputs. The further increase of PE array rows and PE group size cannot achieve 2$\times$ performance gain due to the memory bandwidth limitation. In contrast, when the number of PE array columns is doubled, 2$\times$ speedup can be obtained, since the amount of input loads is unchanged.

\subsection{Comparison with SpinalFlow and GPU}
\label{sec:compare_h2learn}

\begin{figure}[!htbp]
\vspace{-5pt}
\centering
\includegraphics[width=0.48\textwidth]{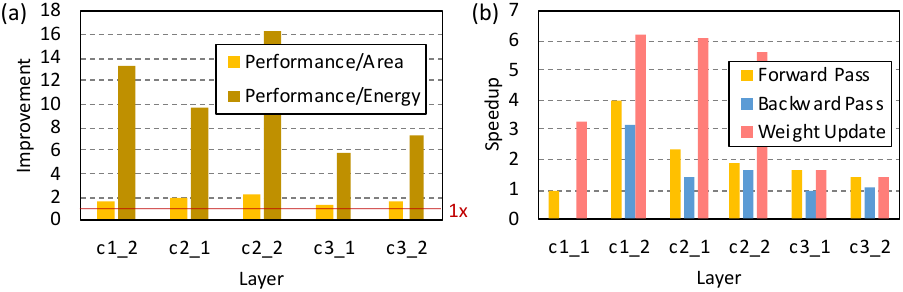}
\caption{Evaluation of \emph{H2Learn} on CIFAR10: (a) Forward Engine compared with SpinalFlow \cite{narayananspinalflow} SNN inference accelerator; (b) throughput of engines compared with NVIDIA V100 GPU.}
\label{fig:result_spinalflow}
\vspace{-5pt}
\end{figure}

We compare \emph{H2Learn} with a state-of-the-art SNN inference accelerator \emph{SpinalFlow} \cite{narayananspinalflow} and NVIDIA V100 GPU \cite{nvidia2017v100} on CIFAR10. The input and output sparsity of the networks are shown in Table \ref{tab:sparsity}. Since \emph{SpinalFlow} only supports inference, in Fig. \ref{fig:result_spinalflow}(a), we compare our Forward Engine with it. We find that \emph{H2Learn} achieves improvement in terms of area and energy. Although \emph{SpinalFlow} skips the computations with zero inputs, they need to store entire weights of all output channels to perform computations associated with a valid input, which consumes large area for weight storage and significant power consumption for data accesses. 
From the functionality perspective, \emph{H2Learn} shows three distinctive characteristics: 1. \emph{H2Learn} targets learning while \emph{SpinalFlow} focuses on inference; 2. \emph{H2Learn} does not have restrictions on coding schemes, while \emph{SpinalFlow} only supports temporal coding; 3. \emph{H2Learn} can support the first encoding layer with hybrid data formats while \emph{SpinalFlow} cannot.

Then, we compare the throughput of engines in \emph{H2Learn} with NVIDIA V100 GPU in Fig.\ref{fig:result_spinalflow}(b). 
We implement the GPU version of SNN learning in Pytorch. Different from the sub-batch-wise pipeline in H2Learn as Fig.\ref{fig:pipeline}, the forward pass and backward pass (along with weight update) are performed sequentially at the grain of the whole batch on GPUs as common handling. We find that \emph{H2Learn} achieves speedup especially in early layers during weight update. In shallow layers, the FM sizes are larger but the number of channels is smaller; besides, weight update needs a 4D rather than 2D Conv. GPU might be inefficient to handle these situations. 

\begin{figure}[!htbp]
\centering
\includegraphics[width=0.485\textwidth]{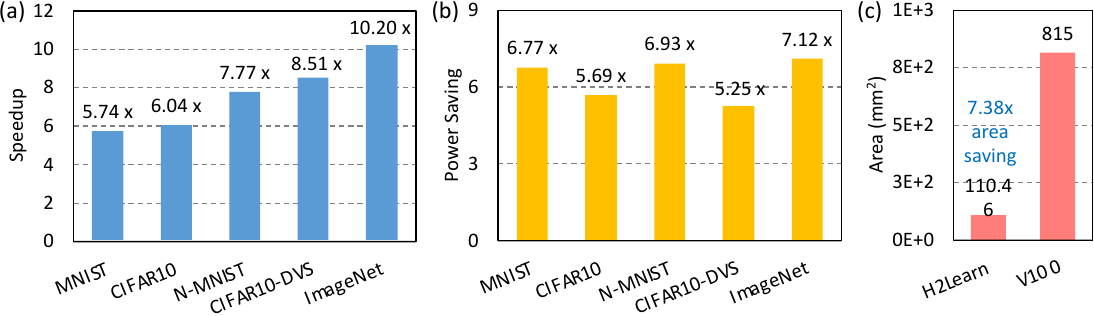}
\caption{Comparison with NVIDIA V100 GPU in terms of (a) throughput, (b) power, and (c) area.}
\label{fig:result_v100}
\vspace{-10pt}
\end{figure}

Fig. \ref{fig:result_v100} shows the comparison between \emph{H2Learn} and NVIDIA V100 GPU during training. Because we focus on the processor design and do not estimate the power of the off-chip memory, here we exclude the HBM power of GPU for fairness. Among the results, \emph{H2Learn} achieve 5.74-10.20$\times$ speedup and 5.25-7.12$\times$ energy saving. We find that \emph{H2Learn} takes more benefits on the ImageNet dataset. The potential reason might be caused by more data preprocessing on GPUs under a large FM size. At last, \emph{H2Learn} is 7.38$\times$ more efficient in area overhead.


\vspace{-5pt}

%% file: text/related.tex
\section{Related Work}

\subsection{Chips for SNN Inference}

Many neuromorphic chips target SNN inference. The early ones adopt mixed-analog-digital circuits based designs \cite{benjamin2014neurogrid, moradi2017scalable} that are usually power efficient but suffer low accuracy and poor programmability. The modern neuromorphic chips prefer fully digital designs \cite{merolla2014million, pei2019towards, deng2020tianjic, narayananspinalflow}. In particular, TrueNorth \cite{merolla2014million} achieves low power via event-driven asynchronous circuits; Tianjic \cite{pei2019towards, deng2020tianjic} bridges ANNs and SNNs using a hybrid architecture with a unified routing infrastructure; Spinalflow \cite{narayananspinalflow} designs an accelerator that can skip redundant computations via input scattering. Different from them for SNN inference, \emph{H2Learn} targets SNN learning.

\subsection{Chips for SNN Learning}
Most of SNN learning chips are designed to implement local synaptic plasticity rules. Similarly, there are also early analog circuits based designs \cite{schemmel2010wafer,qiao2015reconfigurable} and modern digital solutions \cite{frenkel20180,frenkel2019morphic,davies2018loihi,baek2019flexlearn}. Specifically, ODIN \cite{frenkel20180} is the digital version of ROLLS \cite{qiao2015reconfigurable} with only one core per chip, and MorphIC \cite{frenkel2019morphic} is an enhanced version with a hierarchical routing topology; Loihi \cite{davies2018loihi} adopts a many-core architecture, while FlexLearn \cite{baek2019flexlearn} further extends the scope of synaptic plasticity rules. Some studies exploit either SNN inference or training on FPGA \cite{khodamoradi2021s2n2, heidarpur2019cordic, glackin2005novel}. Unlike implementing the local synaptic plasticity rules with lower accuracy, \emph{H2Learn} selects the BPTT learning rule to achieve high accuracy and elaborates the architecture to achieve high efficiency. We also notice a recent work \cite{kulkarni2020chip} supporting BP (not BPTT) for SNNs, but it adopts a LIF variant without temporal propagation, focuses on exploiting the non-volatile memory technology, and only shows results on the small MNIST dataset with two FC layers.

%% file: text/conclusion.tex
\section{Conclusion}

We propose \emph{H2Learn}, an end-to-end accelerator that can implement BPTT-based SNN learning for both high accuracy and high efficiency. Our LUT-based PE design in Forward Engine and Weight Update Engine exploits the spike-based computation; our dual-sparsity-aware Backward Engine exploits both input and output sparsity. 
Compared with the modern NVIDIA V100 GPU, \emph{H2Learn} demonstrates 7.38$\times$ area saving, 5.74-10.20$\times$ speedup, and 5.25-7.12$\times$ power saving on several typical benchmark datasets. 